\documentclass[twocolumn, 9pt]{extarticle}
\usepackage{amsmath,amsthm,amssymb,nicefrac,booktabs}
\usepackage{mathtools}
\usepackage{lineno,hyperref}
\usepackage[table,x11names,dvipsnames,table]{xcolor}
\usepackage{authblk}
\usepackage{subcaption,booktabs}
\usepackage{graphicx}
\usepackage{multirow}
\usepackage[nolist,nohyperlinks]{acronym}
\usepackage[numbers,sort&compress]{natbib}
\usepackage{tabularx}
\usepackage{titlesec}
\usepackage{url}
\usepackage{xurl}
\usepackage{amsfonts}       
\usepackage{nicefrac}       
\usepackage{microtype}      
\usepackage{lipsum}
\usepackage{algorithm}
\usepackage{algorithmic}
\usepackage{multicol}
\usepackage{adjustbox}
\usepackage{wrapfig}
\usepackage{float}
\usepackage[hang,flushmargin]{footmisc} 
\usepackage[group-separator={,},output-decimal-marker={.}]{siunitx}
\usepackage{geometry}
\titlespacing*{\section}
{0pt}{2ex plus 1ex minus 1ex}{2ex plus 1ex}
\titlespacing*{\subsection}
{0pt}{.5ex plus .25ex minus .25ex}{.5ex plus .25ex}

\makeatletter
\renewcommand{\citet}[1]{%
  \def\natexlab##1{}%
  \citeauthor{#1} (\citeyear{#1})~[\citenum{#1}]%
}

\makeatother

\DeclareCaptionLabelFormat{bold}{\textbf{(#2)}}
\graphicspath{{arxiv_submission_13022025_figures/}}  % images default path

\makeatletter
\def\@maketitle{
\raggedright
\newpage
  \noindent
  \vspace{0cm}
  \let \footnote \thanks
    {\hskip -0.4em \huge \textbf{{\@title}} \par}
    \vskip 1.5em
    {\large
      \lineskip .5em
      \begin{tabular}[t]{l}
      \raggedright
        \@author
      \end{tabular}\par}
    \vskip 1em
  \par
  \vskip 1.5em
  }
\makeatother

\newcommand{\algln}{Loss Adapted Plasticity}
\newcommand{\algsn}{LAP}
\newcommand{\parahead}[1]{\textbf{#1}}

\newcommand*\rot{\rotatebox{90}}

\def\D{\mathcal{D}}

\def\temp{\text{temp}}
\def\expt{\mathbb{E}}
\def\Pr{p}

\begin{document}

\title{Training Neural Networks on Data Sources with Unknown Reliability}

\author[1, 2]{Alexander Capstick\thanks{alexander.capstick19@imperial.ac.uk}}
\author[1, 2]{Francesca Palermo}
\author[1, 2]{Tianyu Cui}
\author[1, 2, 3]{Payam Barnaghi}

\affil[1]{Department of Brain Sciences, Imperial College London}
\affil[2]{Care Research and Technology Centre, UK Dementia Research Institute}
\affil[3]{Data Research, Innovation and Virtual Environments (DRIVE) Unit, The Great Ormond Street Hospital}

\setcounter{Maxaffil}{0}
\renewcommand\Affilfont{\itshape\small}

\date{}  
\maketitle

\begin{abstract}
When data is generated by multiple sources, conventional training methods update models assuming equal reliability for each source and do not consider their individual data quality. However, in many applications, sources have varied levels of reliability that can have negative effects on the performance of a neural network. A key issue is that often the quality of the data for individual sources is not known during training. Previous methods for training models in the presence of noisy data do not make use of the additional information that the source label can provide. Focusing on supervised learning, we aim to train neural networks on each data source for a number of steps proportional to the source's estimated reliability by using a dynamic re-weighting strategy motivated by likelihood tempering. This way, we allow training on all sources during the warm-up and reduce learning on less reliable sources during the final training stages, when it has been shown that models overfit to noise. We show through diverse experiments that this can significantly improve model performance when trained on mixtures of reliable and unreliable data sources, and maintain performance when models are trained on reliable sources only.
\end{abstract}

\section{Introduction}
\label{sec:introduction}

Data sources can have differing levels of noise, but in many applications, they are merged to form a single dataset.
For example, in healthcare, data sources can refer to different devices, sites, or human labellers, which are often combined~\citep{tomar2013survey, baro2015toward}. 
These sources may not provide the same level of data quality and could contain noisy features, incorrect labelling, or missing values. 
These problems, if not addressed, can have detrimental effects on the performance and robustness of machine learning models trained on the combined data~\citep{zhang2017understanding, arpit2017closer, neyshabur2017exploring}. 

Two possible solutions for this context naturally arise:
Preprocessing can be used to remove out-of-distribution observations from training~\citep{gamberger2000noise, thongkam2008support, delany2012profiling} but this requires the user to define ``out-of-distribution" for the features and labels, and disregarding data assumes that nothing can be learnt from these noisy examples~\citep{wang2018iterative}. 
Secondly, techniques can be applied to train neural networks on noisy data, of which many exist~\citep{han2020survey, song2022learning}. 
However, these methods consider the noise in the data as a whole and do not utilise the information that can be gained from knowing the data source of an observation.

Within federated learning, work has explored the use of noisy data techniques to address differing data quality issues across data sources~\citep{li2022auto}.
However, federated learning approaches are constrained by the assumption that data cannot be transferred between sources, resulting in limited performance when methods are translated to settings where data sharing is permitted.

We consider the supervised learning case, where data is produced by multiple sources that should be producing data from the same distribution, but where some sources are actually producing noisy data at an unknown rate.
Our goal is to train a neural network on data in this setting that minimises error on a non-noisy test set.

\begin{figure*}[ht]
    \centering
    \includegraphics[width=\linewidth]{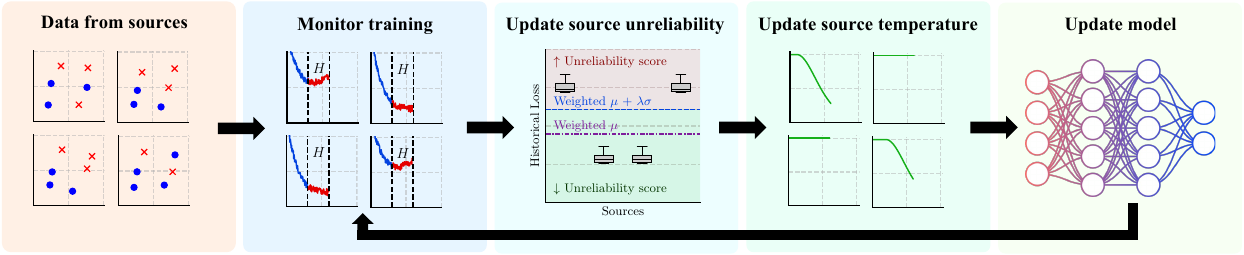}
    \caption{\textbf{Loss Adapted Plasticity:} Here, ER refers to the empirical risk.}
    \label{fig:figure_1}
\end{figure*}

For this, we propose \algln{} (\algsn{}). 
Inspired by the current noisy data literature and likelihood tempering, \algsn{} is a general method for training neural networks on multi-source data with mixed reliability. 
Since we do not know the true noise level of the sources a priori, we maintain a history of the empirical risk on data from each source that we use to temper the likelihood during training.
This temperature is scheduled such that the number of steps in which a model is trained on a source is proportional to its estimated reliability.
An overview of this process is given in Figure \ref{fig:figure_1}.
Hence, a model trained with \algsn{} will benefit from seeing examples from all sources during early training, and reduce learning on less reliable data later in training, when the model is prone to memorising noisy data points~\citep{arpit2017closer, liu2020early, xia2021robust,zhang2017understanding}.
Training a model in this way considerably reduces error on a non-noisy test set when trained on data containing noisy sources.

To illustrate the effectiveness of our method, we present results in various settings and datasets, and compare our method to baselines from the literature on noisy learning \citep{han2018coteaching, wang2021tackling_instance, lixiong2021learning, xia2021robust, wei2022smooth} and federated learning \citep{li2022auto}.

Our work makes the following contributions:
\begin{itemize}
    \item The proposed method, \algsn{}, improves the performance of a neural network trained on data generated from sources with mixed reliability, across a diverse set of experiments in classification, regression, and for multiple data modalities.
    \item When there is no noise within a dataset, \algsn{} maintains performance compared to the standard training method. 
    This means that using our method does not result in less performant models when data does not contain noise.
    \item We provide insights into the mechanics of \algsn{} and how it enables improved performance for models trained with noisy data. 
    As part of this, we present explanations of the hyperparameters introduced and offer intuitions for selecting values.
    \item We provide implementation details and code\footnote{To reproduce our work: \url{https://github.com/alexcapstick/unreliable-sources}; To implement our method on a new task: \url{https://github.com/alexcapstick/loss_adapted_plasticity}} for further development, use in new settings, and ease of reproducibility (Appendix \ref{sec:software}).
\end{itemize}

\section{Related work}

Most of the work on noisy data is focused on noisy \textit{labels} where solutions usually target a combination of four aspects \citep{han2020survey, song2022learning}: sample selection, model architecture, regularisation, or training loss. 
\citet{han2018coteaching} propose ``Co-teaching", a sample selection method in which two neural networks are trained simultaneously. Data for which one model achieves a low training loss are selected to ``teach" the other network, as they are assumed to be more reliable. The authors exploit the fact that neural networks learn clean data patterns and filter noisy instances in early training \citep{zhang2017understanding, arpit2017closer}.
IDPA \citep{wang2021tackling_instance} explores instance-dependent noise, where the noise of the label depends on the characteristics of the observation, in which they estimate the true label of confusing instances using the confidence of the model during training.
The authors \citet{xia2021robust} study how to use regularisation to prevent important weights from being updated late in training, preserving the learning of clean data patterns done in the early training stages.
In contrast, the authors \citet{wei2022smooth} modify the loss during training to study how label smoothing, and negative label smoothing, can improve model accuracy when trained with noisy labels.

Noisy \textit{inputs} are additionally studied in \citet{lixiong2021learning}, where they present RRL, employing two contrastive losses and a noise cleaning method using model confidence during training. This approach requires modifications to the model architecture and a $k$-nearest neighbours search at each epoch.

However, these current works on noisy data do not consider that data can originate from multiple sources, which can have differing noise rates.
We hypothesise that this information is important when data sources contain differing noise rates.

In federated learning, a server trains a global model using local updates on each source \citep{konevcny2016federated}, but when sources are noisy, these algorithms often fail \citep{li2020federated}.
To address the varied data quality, \citet{li2022auto} propose ARFL, which learns global and local weight updates simultaneously. 
The weighted sum of the empirical risk of clients' loss is minimised, where the weights are distributed by comparing each client's empirical risk with the average empirical risk of the best $k$ clients.
The contribution to weight updates from the clients with higher losses is minimised with respect to the updates from other clients -- forcing the global model to learn more from clients that achieve lower losses.
Here, the noise rates of individual data sources are considered during training, however the constraints of the federated learning setting limit this methods use when data can be shared.

Domain generalisation is an area of machine learning studying methods to enable trained models to generalise to unseen data distributions~\citep{wang2022generalizing}.
In this context, many different but related domains are seen during training, with the goal to produce a model that can continue to produce reliable predictions in a new test domain.
To relate this field to our setting, we could consider the data sources in our context as domains, each producing the same data distribution but with different noise levels.
Here, the test domain would produce clean data from the same distribution as the non-noisy data in the training domains.
Since we focus on mitigating the effects of training with noisy data, when viewed through the lens of domain generalisation, our method schedules the training on domains relative to how much information they provide about the test domain.
This is conceptually different from the current literature in domain generalisation but could inspire new research.

\section{Background}

\subsection*{Noisy data}

Data noise can take a variety of forms, and impact both the quality of labels and features.

Noisy labels can be introduced at any point during collection and,
when human experts are involved, are practically inevitable \citep{song2022learning, frenay2014classification, mcnicol2004primer}: An expert may have insufficient information \citep{hickey1996noise, dawid1979maximum};
expert labels are incorrect \citep{hickey1996noise}; the labelling is subjective \citep{marcus1993building}; or there are communication problems \citep{zhu2004class}. 

Further, noisy features can be introduced by poor data processing or measurement errors \citep{lixiong2021learning}.

Noisy data and features can negatively impact the training of a neural network and its performance on a clean test set.
When training a neural network, we use gradient updates to iterate towards an optimal set of model weights.
This is usually done by minimising the empirical risk of model parameters $\theta$ on a dataset of samples $x,y \in \D$ of size $N$:
\begin{equation*}
    R(\D; \theta) = \frac{1}{N} \sum_{i=1}^{N} \mathcal{L} (f_{\theta}(x_i), y_i)
\end{equation*}
Where $R(\D; \theta)$ is the empirical risk and $f_{\theta}$ corresponds to the neural network with parameters $\theta$. 
Updates to the parameters $\theta$ are guided by $\mathcal{L} (f_{\theta}(x_i), y_i)$, which defines the error between a prediction $f_{\theta}(x_i)$ and the true target $y_i$.
When $x_i$ or $y_i$ are noisy, these parameter updates can be misleading, forcing the model to learn relationships that may not exist in the clean data.
Therefore, a model trained with highly noisy data will likely not converge to the same parameters as a model trained with non-noisy data.
This can result in a reduction of accuracy on non-noisy test examples and affect the performance of a model in production.

\subsection*{Noisy data sources}
\label{sec:the_problem_setting}

Since the mechanisms causing noisy data can affect data sources differently, we may observe varied data quality across sources.

In this setting, we have access to a dataset $\D$ that contains observations from multiple mutually exclusive data sources $s \in \mathcal{S}$.
The dataset $\D$ is the union of all data from the sources $s \in \mathcal{S}$:
\begin{equation*}
    \D = \bigcup_{s\in \mathcal{S}} \D_s
\end{equation*}
Since each data source could be separately affected by a noisy data mechanism, this leads to a dataset $\D$ with a total noise rate, $\epsilon$ of:
\begin{equation*}
    \epsilon = \frac{1}{N}\sum_{s \in \mathcal{S}} |\D_s| \times \epsilon_s
\end{equation*}
Here, each source has its own noise rate $\epsilon_s \in [0, 1]$ defining the proportion of the data in that source $\D_s$ that is affected by noise.

\subsection*{Why do we need new methods?}
\label{sec:motivations}

Currently, methods designed to mitigate the effects of noise train a neural network on noisy data to achieve minimal error on a non-noisy test dataset.
We extend this by aiming to achieve minimal error on a non-noisy test dataset, after being trained on \textit{multi-source} data with unknown noise levels.

To see why this could be helpful, consider the $10$ data sources $\{ s_i \}_{i=1}^{10}$ where one source, $s_c$, is producing noisy data with a probability of $0.5$ and all other sources are producing clean data. 
Given a new observation from the noisy source, $x^c \in s_c$, consider the probability that $x^c$ is \textit{noisy}: $\Pr(x^c \notin \mathrm{R})$, where $\mathrm{R}$ is the set of reliable data and assume that we know the noise rate of $s_c$: $\Pr(x \notin \mathrm{R} ~ | ~ x \in s_c)=0.5$.

\textit{Without knowing the data point's source:} The probability that a data point $x$ is unreliable (since $\Pr(x \notin \mathrm{R} ~ | ~ x \notin s_c)=0$) is: $\Pr(x \notin \mathrm{R}) = \Pr(x \in s_c)\Pr(x \notin \mathrm{R} ~ | ~ x \in s_c) =  0.05$.

\textit{When knowing the source of the data point:} The probability that a data point $x^c$ is unreliable ($x^c \in s_c$) is $\Pr(x^c \notin \mathrm{R} ~ | ~ x^c \in s_c) = 0.5$.
Similarly, \mbox{$\Pr(x^{s \neq c} \notin \mathrm{R} ~ | ~x^{s \neq c} \notin s_c) = 0$}.

\begin{figure*}[ht]
    \centering
    \includegraphics[width=0.75\linewidth]{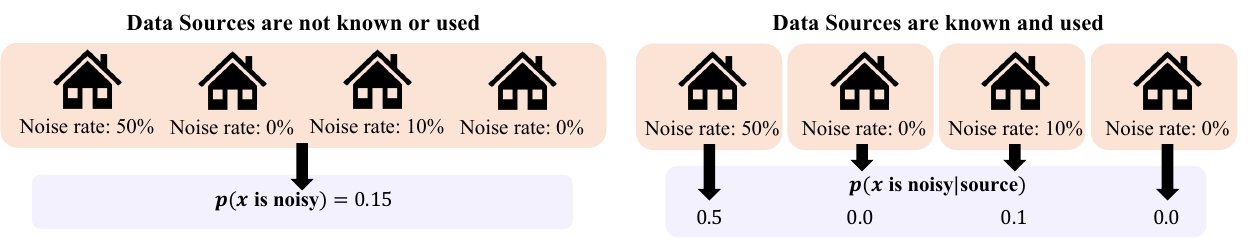}
    \caption{\textbf{Comparing source noise estimation with and without source knowledge.}}
    \label{fig:motivation_for_methods}
\end{figure*}

A similar example is given in Figure \ref{fig:motivation_for_methods}, where we consider multiple noisy sources.
Here, we demonstrate that we can gain more information on the probability of a given sample being noisy by conditioning it on the source it originates from.
Therefore, we hypothesise that this additional information is useful for model training when some data sources are noisy.

\section{Methods}
\label{sec:methods}

Inspired by Co-teaching, ARFL, and loss tempering, we propose \algln{} (\algsn{}), illustrated in Figure \ref{fig:figure_1}.
We continually calculate a reliability score for each source as a function of their historical empirical risks on the training data. This is used to re-weight their negative log-likelihood during training such that the more reliable a source is, the more influence it has when training.
We take advantage of the fact that during training, neural networks learn general data patterns before noisy data and are observed to fit to noisy examples only after substantial progress is made in fitting to clean examples~\citep{zhang2017understanding, arpit2017closer, han2018coteaching, arazo2019unsupervised, yu2019does, shen2019learning}. For more details on this, see Appendix \ref{sec:loss_assumptions}.

In this section, we will use the following shorthand notation: $\Pr_{\theta} ( \cdot ) = \Pr( \cdot | \theta)$.

\subsection*{Tempered likelihood}

Within the literature, tempered likelihood is often used to represent our belief about the aleatoric uncertainty of a dataset~\cite{kapoor2022uncertainty}.
A temperature parameter, applied to the likelihood over samples from a dataset, down-weight their contributions in the calculation of the tempered posterior compared to the prior.
As discussed in \citet{kapoor2022uncertainty}, this can lead to improved estimates of the uncertainty over the posterior parameters of a model and to improved accuracy.
In our work, we use these ideas to temper each source independently.
Therefore, these temperatures represent our belief in the relative uncertainty of the data from each data source.
This tempered likelihood re-weights the contributions of each source to the model training.

With a dataset $\D$, made up of data $\D_s$ collected from sources $s \in S$, and a temperature parameter for each source $T_s$, the tempered log-likelihood for a model parametrised by $\theta$ is:
\begin{equation}
    \log \Pr_{\temp, \theta} (\D)  
    =  \sum_{s \in S} T_s \log \Pr_{\theta} (\D_s)
\end{equation}
However, since a priori $T_s$ is unknown and we do not have access to the noise levels of each source, our prior belief is that $T_s = 1$. For this particular choice of temperatures, the tempered log-likelihood equals the non-tempered log-likelihood $\log \Pr_{ \theta} (\D)$. Therefore, when training starts, we begin with an unmodified optimisation method.

An optimal model in our setting would achieve a maximum log-likelihood on the \textit{clean} data.
Since our sources contain a mix of noisy and non-noisy data points, we aim to approximate this.
Defining $\D_R$ as all clean data in $\D$, we could set the temperature of all noisy sources to $T_s \rightarrow 0$, eliminating their contribution to the overall tempered likelihood. By calculating the weighted mean of the log-likelihood by source, we can arrive at the following approximation:
\begin{equation}
    \max_{\theta} \frac{1}{|S|} \log \Pr_{ \theta} (\D_R) \approx 
    \max_{\theta} \frac{1}{\sum_{s \in S} T_s} \sum_{s \in S} T_s \log \Pr_{\theta} (\D_s \cap \D_R)
\end{equation}
This is an approximation since some sources will contain both noisy and clean samples, and so the approximation would improve as the noise levels of the sources containing noisy examples increases and the number of non-noisy sources increases.

In practice, most sources will contain both clean and noisy examples to varying degrees. Our approach, therefore, detailed in the next section, iteratively updates $T_s$, allowing automatic scheduling of the temperature value.

\subsection*{Source reliability estimation}

To schedule the temperature value during training, we use the historical empirical risk of the model on the batches. The more frequently a data source produces empirical risk values greater than the mean of all other sources, the smaller $T_s$ becomes, and the less influence that source has during training. Under the dynamics that neural networks learn clean data before noisy data, this allows us to take advantage of the approximation given in the previous section during late training, and learn from clean data within noisy sources earlier in model training.

To provide greater control over this process, we set $T_s = f(C_s)$, where $f: [0, +\infty) \rightarrow (0, 1]$ is a positive and monotonically decreasing function. This allows us to generalise the method to other scheduling functions and interpret the value of $C_s$ as a source's perceived unreliability.

We update $C_s$ by considering the number of times that a given data source provides data that achieves a significantly greater negative log-likelihood (NLL) than the average of the other data sources.
To do this, we maintain a history (of length $H$) of NLLs for each of the batches. When updating the value $C_s$ for a single source, we consider the weighted mean $\mu_{s'}$ and standard deviation $\sigma_{s'}$ of the NLL values of all other sources from their stored history.
If the mean NLL value over the history for the source in question $\mu_s$ is greater than $\mu_{s'} + \lambda \sigma_{s'}$ for a chosen $\lambda$ (hyperparameter $\lambda>0$) then we increase $C_s$, otherwise we reduce it:
\begin{align}
    \label{eq:lap_method_equation}
    \begin{array}{c}
        C_{s} = 
            \begin{cases} 
              C_s + 1, & \mu_s > \mu_{s'} + \lambda \sigma_{s'} \\
              C_{s} - 1, & \text{otherwise}
           \end{cases}
    \end{array}
\end{align}
To ensure that $C_s$ remains positive and within the domain of $f$, we clip its value below at $0$. This means that $C_s$ corresponds to how much a source's data historically deviates from those patterns learnt by the model.
This translates to only decreasing the source's temperature $C_s$ (and therefore its perceived unreliability), if its log-likelihood is at most $\lambda$ standard deviations less than the expected tempered log-likelihood on all other sources.

The algorithm used to calculate $C_s$ based on a history of losses is given in Algorithm \ref{sec:appendix:alg_lap_mathod}.

\begin{algorithm}
\caption{Calculating $C_s$ at a given step}\label{alg:lap_method}
\begin{algorithmic}
\REQUIRE $\lambda > 0$ : Leniency
\REQUIRE $L \in \mathbb{R}^{S \times H}$ : Source loss history of length $H$
\REQUIRE $C \in \mathbb{R}^{S}$ : The current unreliability
\STATE $L_s = L[s]$ \COMMENT{Loss history on source $s$}
\STATE $C_s = C[s]$ \COMMENT{Unreliabilty for source $s$}
\STATE $\mu_s = \texttt{mean} ( L_s )$
\STATE $W[s'] = f(C[s'])$
\STATE $\mu_{s'} = \texttt{weighted\_mean} ( ~ L[s'], ~\texttt{weights}=W[s'] ~)$
\STATE $\sigma_{s'}^2 = \texttt{weighted\_var} (~ L[s'], ~\texttt{weights}=W[s'] ~)$
\IF{$\mu_s > \mu_{s'} + \lambda \sigma_{s'}$}
    \STATE $C_s = C_s + 1$
\ELSIF{$\mu_s \leq \mu_{s'} + \lambda \sigma_{s'}$}
    \STATE $C_s = \max \{ ~ 0, ~ C_s - 1 ~ \}$
\ENDIF
\end{algorithmic}
\end{algorithm}

In Appendix \ref{sec:appendix_prove_no_noise_same_as_standard}, we show theoretically that when there is no noise in the dataset, and batches are large, our tempered likelihood training uses the same update as standard maximum likelihood training at each parameter step, with a high probability.
This is more likely the larger $\lambda$ is.

This hyperparameter $\lambda$, which we refer to as the \textit{leniency}, relates to the probability of incorrectly reducing the temperature of a non-noisy source when all sources are non-noisy (Appendix \ref{sec:appendix_lambda_desc}).
When a dataset contains noisy sources, we expect the temperature of those noisy sources to be small (i.e.: $f(C_s) << 1$) after a number of training steps corresponding to their relative noise level -- naturally enabling learning from noisy sources for a length of time that reflects their ``usefulness". 

\subsection*{Implementation details}

We require that our temperature, $f(C_s)$, monotonically decreases as $C_s$ increases so that we reduce the impact of a source on the overall tempered likelihood as its perceived unreliability increases. 
We also would like $f(C_s)$ to be robust to small changes in $C_s$ around $0$, to ensure that fluctuations in the log-likelihood between batches do not incorrectly lead to a reduction in the temperature of a source. 
Because of its use as an activation function and so its wide accessibility, we therefore choose to use the following for our experiments:
\begin{equation*}
    f(C_s) = 1-\tanh^2 ( 0.005 \cdot \delta \cdot  C_s ) = 1 - d_s
\end{equation*}
Where $\delta$, which we refer to as the \textit{depression strength}, rescales the value $C_s$ to control the rate of reduction in temperature. In our analysis, we additionally scale $\delta$ by $0.005$ so that its value is more human-readable. We refer to $\tanh^2 ( 0.005 \cdot \delta \cdot  C_s )$ as the depression value $d_s$ for simplicity in later sections. 

We find that this choice of $f$ provides some nice properties: $y = \tanh^2 x$ has gradient of $0$ at $x=0$, and is asymptotic to $y = 1$, which are important qualities for calculating the model update scaling. 
The stationary point at $x=0$ also means that small perturbations in the unreliability of each source around $C_s = 0$ have little effects on the scaling of model updates, which reduces the consequences of randomness in loss values from a given source and ensures that source contributions are only significantly scaled once it is clear that their loss values are consistently higher than those of the other sources. 
Secondly, since $y = \tanh^2 x$ is asymptotic to $y = 1$, the scale factor, $1-d_s$ for sources that are calculated to be unreliable eventually reduces to $0$, allowing the model to ``ignore" these data points as if they were masked in late training. 
Further, the use of $tanh$ as an activation function in other areas of machine learning mean that it is widely accessible.
The value of $0.005$ used within the $\tanh^2$ function allows the depression strength parameter $\delta$ to be on the scale of $1$, as shown in Figure \ref{fig:toy_example}. 
This value is only used to aid interpretation of the hyperparameters and its value could easily be absorbed by $\delta$.
For more detail on design decisions, see Appendix \ref{sec:design_decisions}.

Although we find this choice of $f(C_s)$ to work well in practice, the \algsn{} method is designed so that other functions can be used to update the influence of sources based on their perceived reliability.
The only requirement on $f(C_s)$ is that it is monotonically decreasing and $f: [0, +\infty) \rightarrow (0, 1]$.

\begin{figure*}[ht]
    \centering
    \includegraphics[width=1\linewidth]{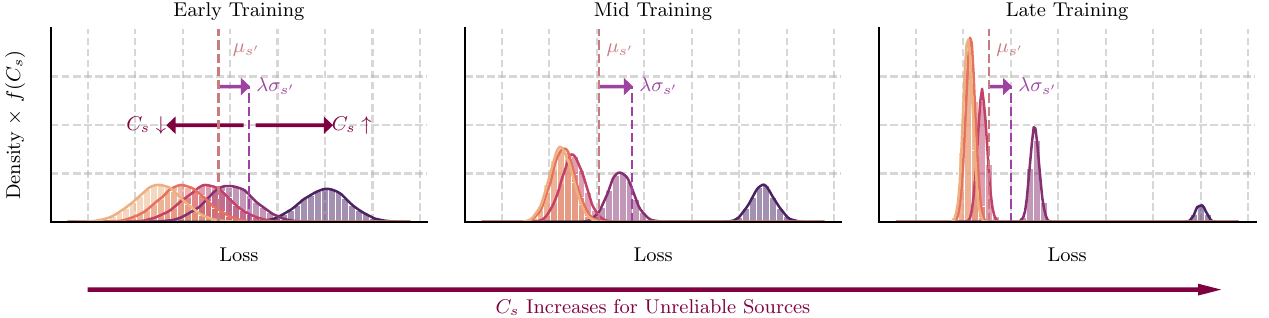}
    \caption{\textbf{Visualisation of Equation \ref{eq:lap_method_equation}.} Each colour represents the loss values from a single source over a small number of steps, with its density weighted by its temperature, $w(s) = f(C_s)$. This shows how sources contribute to $\mu_{s'}$ and $\sigma_{s'}^2$ as their $C_s$ changes during training and given the leniency $\lambda$. These values are synthetic and for demonstration.}
    \label{fig:source_loss_with_Czeta}
\end{figure*}

During training, the gradient contributions from source $s$ are then scaled as follows:
\begin{equation}
    \hat{g_s} = f(C_s) g_s = (1-d_s) g_s 
\label{equation:gradient_depression}
\end{equation}
Where $\hat{g_s}$ is the gradient update contribution from a source $s$.
As $(1-d_s)$ is a scalar, this method can be interpreted as scaling gradient contributions (presented here), loss re-weighting as in Figure \ref{fig:source_loss_with_Czeta}, or likelihood tempering as discussed previously in this section. 
Furthermore, a model evidence interpretation of \algsn{} can be found in Appendix \ref{sec:model_interpretations}.

Note that in this construction, we make no assumptions about the type of observations contained in $\D$. 
Therefore, our method is applicable to images, natural language, time series, and tabular data for both regression and classification tasks.

\section{Results}
\label{sec:results}

\subsection*{Baseline methods}
\label{sec:baseline_methods}

To contextualise our approach, we evaluate varied baseline methods: 
(1) ARFL, designed to tackle label and input noise in a federated learning setting~\citep{li2022auto}; 
(2) IDPA, a probabilistic method for instance dependent label noise~\citep{wang2021tackling_instance} which modifies the training loss; 
(3) ``Co-teaching", which trains two models simultaneously to perform sample selection based on loss values during training~\citep{han2018coteaching}; 
(4) CDR, which regularises model weights such that important parameters, learnt during the early training phase, do not considerably change when the model fits to the noisy data~\citep{xia2021robust};
(5) Label Smoothing, which uses a combination of the true label and a uniformly weighted soft label during training~\citep{wei2022smooth};
(6) RRL, which uses contrastive learning and a $k$ nearest neighbours search to enforce a smoothness constraint on learnt representations~\citep{lixiong2021learning}, modifying model architecture and training loss; 
and (7) an identical model but without any specific modifications for tackling noise. Baseline methods were selected based on the availability of code, their use as baselines in the literature, and for variety in the methods used to evaluate the performance of \algsn{}. 
Although our setting assumes noise is independent of features, we felt it was still beneficial to include IDPA, which is designed for instance dependent label noise. 
Note that because ARFL is a federated learning approach, data cannot be shuffled in the same way as the other models, since each client trains on a single source.

It is important to note that both IDPA and Co-teaching require the training of a model twice, making \algsn{} significantly less computationally expensive.

All baselines are implemented using the available code and trained using the recommended parameters with the model and data we test. For further details, see Appendix \ref{sec:experiment_information}.

\subsection*{Experimental design}
\label{sec:experimental_design}

To evaluate \algsn{}, we employ various techniques to produce noisy data, extending the methods in \citet{li2022auto, wang2021tackling_instance, han2018coteaching, lixiong2021learning} and test on eleven datasets from computer vision, healthcare time-series, natural language processing, and tabular regression for diverse experiments.

Following \citet{li2022auto, wang2021tackling_instance, han2018coteaching, lixiong2021learning}, we test our proposed method and baselines on CIFAR-10, CIFAR-100 \citep{krizhevsky2009learning}, and F-MNIST \citep{xiao2017fashion} forming many comparisons with the literature. Along with Tiny-Imagenet and Imagenet~\citep{deng2009imagenet}, these five datasets form well-studied computer vision tasks, easing reproducibility. 
Additionally, we use a human-labelled version of CIFAR-10, titled CIFAR-10N~\citep{wei2022learning}, for which we use the ``worst labels" allowing us flexibility in our experiments.
We then study an electrocardiograph (ECG) dataset, PTB-XL \citep{Wagner2020}; a time-series classification task with the goal of classifying normal and abnormal cardiac rhythms and for which noise can be simulated following \citet{emiwong2012} and random labelling to understand \algsn{} applied to time-series data with multiple types of noise. 
Additionally, a natural language sentiment analysis task allows us to compare \algsn{} with the literature, for which we use the IMDB data set \citep{maas2011imdb} that contains movie reviews and their sentiment. 
Next, we use the GoEmotions dataset \citep{demszky2020goemotions}, a natural language emotion prediction task, which contains unbalanced source sizes and class distributions in the real world, allowing us to test the robustness of \algsn{} to varied source constructions.
Finally, to illustrate the use of our method for regression, we present results on the California Housing dataset~\citep{pace1997sparse}.
For further information, see Appendices \ref{sec:dataset_further_information} and \ref{sec:source_distribution_goemotions}.

To simulate data sources for CIFAR-10, CIFAR-100, F-MNIST, Imagenet, IMDB, and California Housing, data is uniformly split into $10$ distinct groups, and for Tiny-Imagenet we use $100$ groups to study larger numbers of sources.
For CIFAR-10 and CIFAR-100, $4$ and $2$ of these sources are chosen to be noisy respectively; for F-MNIST, Imagenet, and Tiny-Imagenet $6$, $5$ and $40$ are chosen; and for IMDB and California Housing $4$ are chosen. These are in line with the noise rates used in the literature (often $20\%$, $40\%$, $50\%$).
For CIFAR-10, in Appendix \ref{sec:cifar10_lots_of_sources} we also significantly increase the number of sources.
To generate noise for the vision datasets, we extended the methods in \citet{li2022auto}:
(1) Original Data: No noise is applied to the data; 
(2) Chunk Shuffle: Split features into distinct chunks and shuffle. This is only done on the first and/or second axes of a given input; 
(3) Random Label: Randomly assign a new label from the same code; 
(4) Batch Label Shuffle: For a given batch of features, randomly shuffle the labels; 
(5) Batch Label Flip: For a given batch of data, assign all features in this batch a label randomly chosen from the same batch; 
(6) Added Noise: Add Gaussian noise to the features, with mean $=0$ and standard deviation $=1$; 
(7) Replace with Noise: Replace features with Gaussian noise, with mean $=0$ and standard deviation $=1$.

For IMDB, GoEmotions, and California Housing, we use random labelling.

PTB-XL was labelled by $12$ nurses, naturally forming data sources. However, since this data is high quality, synthetic noise is required. We add Gaussian noise to sources' ECG recordings (simulating electromagnetic interference as in \citet{emiwong2012}) and label flipping to simulate human error in labelling. This also allows us to test the setting with multiple noise types.
Here, data from sources are upsampled so that each source contains the same number of observations. For experiments with PTB-XL, we linearly increase the number of noisy sources from $1$ to $8$ (out of $12$ in total), and for each number of noisy sources we set the noise level for each source linearly from $0.25$ to $1.0$. For example, when training with $4$ noisy sources, sources haves noise levels of $0.25$, $0.50$, $0.75$, and $1.0$.

Since GoEmotions is annotated by $82$ raters, their identification number is used to form the data sources. Some raters contributed a handful of data points, whilst others labelled thousands; with each rater providing a different distribution of class labels. This allowed us to explore the real-world case of data with sources of imbalanced size and label distribution. A detailed discussion of the imbalanced sources is given in Appendix \ref{sec:source_distribution_goemotions}.

Although CIFAR-10N contains real-world noise, we still have to split the data into sources. For each experiment, we assign sources to evaluate the varied levels of noise and the number of noisy sources. As in PTB-XL, we linearly increase the number of noisy sources from $1$ to $7$ (out of $10$ in total), and for each set of noisy sources, we linearly increase the noise level from $0.25$ to $1.0$.

For this, seven base model architectures (of multiple sizes) are evaluated: A Multilayer Perceptron, Convolutional Neural Networks, a 1D and 2D ResNet \citep{he2016deep}, an LSTM \citep{hochreiter1997long}, and a transformer encoder \citep{vaswani2017attention}  (Appendix \ref{sec:experiment_information}).

In all experiments, data points contain an observation, source, and target, which are assigned to mini-batches in the ordinary way. Features and labels are passed to the model for training, whilst sources are used by \algsn{} (Appendix \ref{sec:experiment_information}). The test sets contain only clean observations.

\subsection*{Intuition for the hyperparameters}

Figure \ref{fig:source_loss_with_Czeta} shows how the training process evolves over time when using \algsn{}. At first, all sources are considered equally reliable (that is, $T_s = 1$), and therefore the weighted mean of the loss values is the mean of all sources. As the source temperature changes, the weighed mean plus $\lambda$ standard deviations of the losses moves toward the sources with lower expected losses, allowing more learning from these over time.

\begin{figure*}[ht]
    \centering
    \includegraphics[width=1\linewidth]{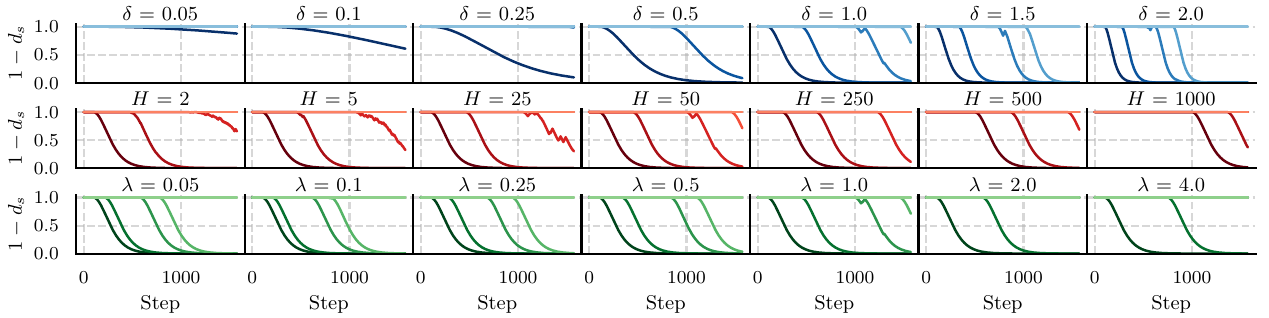}
    \caption{\textbf{Effect of the introduced parameters on training.} Section \ref{sec:methods}, introduces three parameters that control the effects of \algsn{}. $1-d_s$ is multiplied by the gradient (equivalently, loss) contribution from a given source before the model is updated. Here, we show these values for each source (the different coloured lines) during model training on synthetic data (Appendix \ref{sec:toy_example}). Unless stated in the title of a given plot, the parameters of \algsn{} were set to $H=25$, $\delta=1.0$, $\lambda=1.0$. We had $5$ sources with noise levels of $0.0$, $0.025$, $0.05$, $0.25$, and $1.0$ (a darker colour indicates a higher noise rate).}
    \label{fig:toy_example}
\end{figure*}

Figure \ref{fig:toy_example} illustrates how $1-d_s$ (the temperature) in Equation \ref{equation:gradient_depression} develops during training for each source and different values of $H$, $\delta$, and $\lambda$. The first row shows how varying the depression strength, $\delta$ affects the rate at which the learning on a source is reduced. The second row shows how the history length, $H$ influences $1-d_s$, where we can see that a larger value allows for a ``smoother" transition. With the exception of small or large values, this parameter has the least effect on training. The third row illustrates how the leniency $\lambda$ in Equation \ref{eq:lap_method_equation} and Algorithm \ref{alg:lap_method} affects training. Small values of $\lambda$ reduce the influence of noisy sources earlier, whilst larger values miss some noisy sources altogether. Here, the sources are coloured by their noise level, demonstrating that \algsn{} reduced the influence of a source on training in a number of steps corresponding to the noise level of the source.

In our experiments, the parameters of \algsn{} were set at $H=25$, $\delta=1.0$, $\lambda=0.8$ unless otherwise stated in Appendix \ref{sec:experiment_information}. These were chosen based on the synthetic experiments presented in Figure \ref{fig:toy_example} and the discussion of the effects of these hyperparameters provided in Appendix \ref{sec:hparam_discussion}.

\subsection*{\algsn{} and the baselines}

\begin{table*}[ht]
\begin{center}
\caption{\textbf{Comparison of \algsn{} with the baselines.} Mean ± standard deviation of the maximum test accuracy (\%) of $5$ repeats of the baselines and \algsn{} on synthetic data with different noisy types. For CIFAR-100 these numbers represent the top 5 accuracy. For CIFAR-10, CIFAR-100, and F-MNIST, the number of noisy sources are $4$, $2$, and $6$ out of $10$ respectively. Unreliable sources are each 100\% noisy. All values in bold are within 1 standard deviation of the maximum score. Note that IDPA and Co-teaching require almost twice the training time in comparison to Standard, ARFL, CDR, Label Smoothing (LS), and \algsn{}.}
\label{table:synthetic_results}
\scriptsize
\begin{adjustbox}{max width=1.75\textwidth,center}
\begin{tabular}{ccccccccc}
\cmidrule(lr){3-9}
 & & \multicolumn{7}{c}{Model Types} \\
\toprule
 & Noise Type & Standard & ARFL (\citeyear{li2022auto}) & IDPA (\citeyear{wang2021tackling_instance}) & Co-teaching (\citeyear{han2018coteaching}) & CDR (\citeyear{xia2021robust}) & LS (\citeyear{wei2022smooth}) & \algsn{} (Ours) \\
\midrule
\multirow{7}{*}{\rot{CIFAR-10}} & Original Data & 77.82 ± 0.37 & 74.89 ± 1.67 & \textbf{79.89 ± 1.01} & \textbf{80.04 ± 0.49} & 77.2 ± 1.59 & 78.47 ± 0.72 & 77.52 ± 0.9 \\
 & Chunk Shuffle & 72.12 ± 0.95 & 68.42 ± 0.71 & 72.68 ± 0.96 & \textbf{74.77 ± 0.3} & 70.02 ± 1.35 & 70.1 ± 0.97 & 74.29 ± 1.43 \\
 & Random Label & 67.84 ± 1.84 & 69.93 ± 1.88 & 55.66 ± 1.54 & 68.61 ± 1.78 & 65.61 ± 0.97 & 67.92 ± 2.66 & \textbf{73.44 ± 0.9} \\
 & Batch Label Shuffle & 68.75 ± 1.36 & 67.12 ± 1.93 & 68.19 ± 1.48 & 71.68 ± 0.92 & 68.21 ± 1.06 & 67.35 ± 1.23 & \textbf{73.94 ± 0.33} \\
 & Batch Label Flip & 67.12 ± 0.7 & 66.16 ± 1.96 & 69.65 ± 1.54 & 71.18 ± 0.76 & 66.59 ± 1.79 & 68.39 ± 2.14 & \textbf{73.85 ± 0.92} \\
 & Added Noise & 70.23 ± 0.76 & 68.59 ± 1.38 & 70.94 ± 1.61 & \textbf{72.29 ± 0.61} & 68.3 ± 1.32 & 69.48 ± 1.33 & \textbf{72.63 ± 1.24} \\
 & Replace With Noise & 72.79 ± 0.67 & 67.01 ± 1.02 & 71.96 ± 1.53 & \textbf{73.87 ± 0.57} & 71.02 ± 1.04 & 72.97 ± 0.83 & \textbf{73.84 ± 0.55} \\
 \midrule
\multirow{7}{*}{\rot{CIFAR-100}} & Original Data & 75.95 ± 1.01 & 60.42 ± 1.91 & \textbf{77.78 ± 0.95} & 76.25 ± 0.54 & 75.91 ± 1.11 & 75.42 ± 0.49 & 76.01 ± 0.54 \\
 & Chunk Shuffle & 68.86 ± 0.74 & 56.76 ± 2.69 & 66.82 ± 0.85 & \textbf{69.58 ± 0.51} & 68.65 ± 0.91 & 68.15 ± 0.95 & \textbf{70.41 ± 0.96} \\
 & Random Label & 58.34 ± 0.91 & 48.85 ± 3.2 & 49.5 ± 1.03 & 61.2 ± 0.54 & 57.12 ± 0.76 & 58.3 ± 0.93 & \textbf{69.05 ± 0.38} \\
 & Batch Label Shuffle & 64.8 ± 0.97 & 58.2 ± 4.45 & 64.34 ± 2.09 & \textbf{69.34 ± 0.69} & 65.69 ± 1.15 & 64.73 ± 1.12 & \textbf{69.31 ± 1.4} \\
 & Batch Label Flip & 61.89 ± 0.97 & 56.51 ± 3.36 & 64.61 ± 2.1 & \textbf{69.04 ± 1.19} & 60.01 ± 1.16 & 60.64 ± 2.05 & \textbf{69.82 ± 0.78} \\
 & Added Noise & 65.28 ± 0.77 & 58.35 ± 4.16 & 64.44 ± 1.11 & 66.02 ± 0.36 & 65.24 ± 1.32 & 65.38 ± 1.28 & \textbf{67.69 ± 1.17} \\
 & Replace With Noise & \textbf{67.76 ± 1.33} & 58.88 ± 4.33 & 66.06 ± 0.42 & \textbf{68.52 ± 1.31} & \textbf{68.48 ± 1.31} & \textbf{67.58 ± 0.92} & \textbf{68.71 ± 1.22} \\
 \midrule
\multirow{7}{*}{\rot{F-MNIST}} & Original Data & \textbf{83.64 ± 0.33} & 82.0 ± 0.4 & \textbf{83.69 ± 0.65} & 79.09 ± 1.14 & 82.42 ± 2.27 & \textbf{83.74 ± 0.27} & \textbf{83.52 ± 0.21} \\
 & Chunk Shuffle & 77.12 ± 1.79 & 77.69 ± 0.51 & 77.45 ± 2.21 & 74.74 ± 1.21 & 76.45 ± 1.7 & 78.17 ± 0.75 & \textbf{82.18 ± 1.26} \\
 & Random Label & 76.22 ± 8.06 & 77.31 ± 3.82 & 76.57 ± 7.16 & 77.41 ± 4.47 & 75.33 ± 7.27 & 74.79 ± 4.84 & \textbf{81.8 ± 1.37} \\
 & Batch Label Shuffle & \textbf{82.65 ± 0.13} & 79.11 ± 1.71 & \textbf{82.82 ± 0.43} & 82.3 ± 0.32 & \textbf{82.88 ± 0.44} & \textbf{82.63 ± 0.45} & \textbf{82.45 ± 0.38} \\
 & Batch Label Flip & 80.25 ± 1.87 & 78.85 ± 1.78 & \textbf{82.31 ± 0.27} & 80.43 ± 0.81 & 81.29 ± 0.58 & 80.56 ± 0.67 & 79.87 ± 0.7 \\
 & Added Noise & 77.88 ± 1.86 & 73.28 ± 1.31 & \textbf{78.42 ± 2.1} & 75.72 ± 1.55 & 76.21 ± 2.37 & \textbf{79.02 ± 0.75} & \textbf{78.72 ± 2.02} \\
 & Replace With Noise & 79.77 ± 0.8 & 74.12 ± 1.36 & 80.64 ± 0.42 & 78.76 ± 1.06 & 78.24 ± 0.93 & 79.94 ± 0.64 & \textbf{82.94 ± 0.31} \\
\bottomrule
\end{tabular}
\end{adjustbox}
\end{center}
\end{table*}

Table \ref{table:synthetic_results} shows the results of the baselines and \algsn{} on CIFAR-10, CIFAR-100 and F-MNIST with varied noise. 
This illustrates that in a high noise setting, \algsn{} often allows for improved test accuracy over the baselines, four different approaches to learn from noisy labels (IDPA, Co-teaching, CDR, and Label Smoothing), an approach from federated learning (ARFL), and training a model in the standard way.
Interestingly, IDPA and Co-teaching achieve a higher accuracy on the original data in CIFAR-10 and 100, likely because IDPA and Co-teaching require the training of a model for twice as long or twice, \textit{almost doubling training time} over \algsn{}, ARFL and standard training.

On CIFAR-10 and 100, the performance improvement with \algsn{} is most apparent with the random labelling noise. For example, on CIFAR-100, \algsn{} achieves a $+18.4\%$ increase in top-5 accuracy over standard training, whilst IDPA performs worse. Although the performance improvements of \algsn{} are less dramatic on F-MNIST due to the simpler nature of the dataset, the model still demonstrates better accuracy in certain noise conditions, particularly with chunk shuffling and random label noise.

In these experiments, \algsn{} achieves the best accuracy or is within a few percentage points of the best accuracy. In the latter case, the best accuracy is then often achieved by a method that requires longer training but that is less robust to other noise types. Therefore, \algsn{} overall achieves consistently higher accuracy while requiring less computation.

To further illustrate the potential accuracy improvement from using \algsn{}, in Appendix \ref{sec:synthetic_results_percentage_difference} we also present these results as a percentage difference from the standard training method. In the following more challenging experiments, the improvement in the accuracy of using \algsn{} becomes clearer. Furthermore, although IDPA and co-teaching are limited to classification tasks, \algsn{} also works on regression (Appendix \ref{sec:california_housing}).

\subsection*{\algsn{} in conjunction with RRL} 

Since RRL requires significant changes to the model architecture and data augmentation, we separately test the large architecture (Appendix \ref{sec:experiment_information}) and the experimental setup used in \citet{lixiong2021learning} for a better comparison. Here, we test RRL with and without \algsn{} to evaluate its ability to be used in conjunction with other methods. Table \ref{table:synthetic_presnet_results} presents these results on CIFAR-10, and further highlights the use of \algsn{} to improve model performance on data generated from multiple sources. 
Here, \algsn{} supplements the accuracy obtained by RRL, which uses a contrastive learning approach to tackle noisy features and labels that is improved upon by further applying our method to make use of the source values.
We observe that \algsn{} increases all metric scores except for ``Replace With Noise", where the difference is relatively small, and ``Original Data", where both methods perform equally. 
Again, the most substantial increase in accuracy comes from random labelling noise, where \algsn{} improves RRL by approximately $+5\%$.
These results are supplemented by Appendix \ref{sec:cifar_different_noise_results}, where we test different numbers of noisy sources and noise rates.

%\begin{wraptable}{R}{8cm}
\begin{table}[ht]
\begin{center}
\caption{\textbf{RRL and RRL + \algsn{} results.} Mean ± standard deviation of the maximum test accuracy (\%) of $10$ repeats of RRL and \algsn{} on CIFAR-10 data with different types of noise. Here, $4$ out of $10$ sources are 100\% noisy.}
\label{table:synthetic_presnet_results}
\scriptsize
\begin{tabular}{ccc}
\cmidrule(lr){2-3}
 & \multicolumn{2}{c}{Model Types} \\
\toprule
Noise Type & RRL (\citeyear{lixiong2021learning}) & RRL + \algsn{} (Ours) \\
\midrule
Original Data & \textbf{87.67 ± 0.37} & \textbf{87.54 ± 0.22} \\
Chunk Shuffle & 82.93 ± 0.29 & \textbf{84.27 ± 0.31} \\
Random Label & 76.04 ± 1.43 & \textbf{80.31 ± 0.58} \\
Batch Label Shuffle & 77.66 ± 0.71 & \textbf{80.84 ± 0.51} \\
Batch Label Flip & 78.81 ± 0.66 & \textbf{82.02 ± 0.45} \\
Added Noise & 78.51 ± 0.74 & \textbf{81.70 ± 0.48} \\
Replace With Noise & \textbf{80.05 ± 0.65} & 79.00 ± 0.75 \\
\bottomrule
\end{tabular}
\end{center}
\end{table}
%\end{wraptable}

\subsection*{Varied noise level and source sizes with real-world time-series data} 

\begin{figure*}[ht]
     \centering
     \begin{subfigure}[b]{0.49\textwidth}
         \centering
         \includegraphics[width=\textwidth]{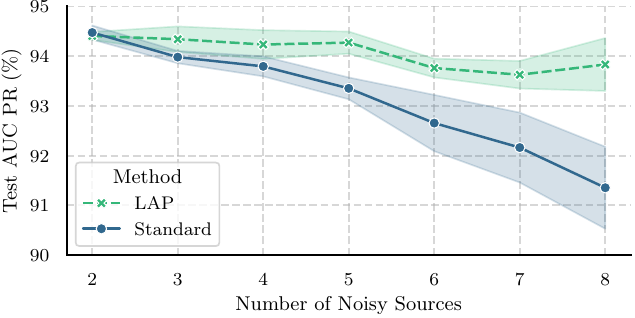}
         \caption{\textbf{AUC precision-recall on PTB-XL.}}
         \label{fig:ptbxl_aucpr_vs_corruption}
     \end{subfigure}
     \begin{subfigure}[b]{0.49\textwidth}
         \centering
         \includegraphics[width=\textwidth]{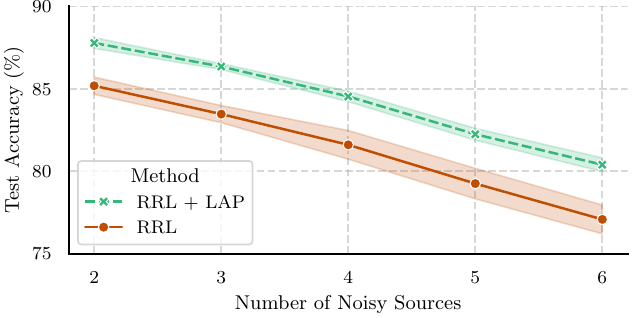}
         \caption{\textbf{Accuracy on CIFAR-10N.}}
         \label{fig:cifar10n_acc_vs_corruption_presnet}
     \end{subfigure}
    \caption{\textbf{\algsn{} results with a varied number of sources and noise levels.} In \subref{fig:ptbxl_aucpr_vs_corruption} we show the area under the precision-recall curve for standard training and using \algsn{} on PTB-XL with label noise and simulated ECG interference noise for $12$ total sources. In \subref{fig:cifar10n_acc_vs_corruption_presnet} we show the accuracy on CIFAR-10N with real human labelling noise when using RRL and RRL + \algsn{}, with $10$ sources. In both, the noise of the sources varies linearly from $25\%$ to $100\%$ for each number of noisy sources. The lines and error bands represent the mean and standard deviation of the maximum value for each of the 5 repeats. These figures illustrate that \algsn{} maintains higher performance as noise rates increase.}
    \label{fig:varied_number_of_sources}
\end{figure*}

Figure \ref{fig:ptbxl_aucpr_vs_corruption} compares the results of not using and using \algsn{} for a time-series classification task on normal and abnormal cardiac rhythms with both label and input noise. To evaluate the model on data sources with differing noise levels, for each number of noisy sources we set the noise levels of the sources in linear increments between $25\%$ and $100\%$. Additionally, we linearly increase the number of these noisy sources and measure the area under the precision-recall curve (AUC PR) on the test set. These results show that \algsn{} allows for improved model training on data with sources of varied noise and with multiple noise types.
Figure \ref{fig:ptbxl_aucpr_vs_corruption} also illustrates that in this setting, \algsn{} is robust to increases in noise within the dataset, as AUC PR does not degrade significantly, especially when compared to not using \algsn{}. Furthermore, the standard deviation of the results is smaller, suggesting that \algsn{} is more consistent in its AUC-PR.
When only two sources contain noise, the total noise rate in the dataset is only $10\%$, which has little negative effect on standard training.

\subsection*{Real-world noisy labels} 

To further demonstrate the effectiveness of \algsn{} in conjunction with other noisy data methods and on real-world noise, we now present the results on a dataset with human labelling of varied noise rates. Figure \ref{fig:cifar10n_acc_vs_corruption_presnet} illustrates the results on CIFAR-10N, a noisy human-labelled version of CIFAR-10, where we observe a pattern similar to that depicted in Figure \ref{fig:ptbxl_aucpr_vs_corruption}. 
In both cases, \algsn{} improved the accuracy on noisy data at all noise levels tested, and the standard deviation of the accuracy is reduced. 
However, we also note that in this case the accuracy of \algsn{} degraded faster than the AUC PR in Figure \ref{fig:ptbxl_aucpr_vs_corruption} as the number of noisy sources increased.
This could be for various reasons, such as differing noise types, model baselines, and data type.
In Appendix \ref{sec:cifar10n_small_network} we also show the results of this experiment when using the same model architecture as in Table \ref{table:synthetic_results}, and with standard training as a baseline, instead of RRL, where we arrive at the same conclusions.

\subsection*{\algsn{} on a noisy natural language task}

%\begin{wraptable}{R}{11cm}
\begin{table*}[ht]
\begin{center}
\caption{\textbf{Baselines and \algsn{} results on a natural language task.} Mean ± standard deviation of maximum test accuracy (\%) of $5$ repeats of the methods on the IMDB dataset with different types of noise. Here, $4$ out of $10$ sources are 100\% noisy.}
\label{table:imdb_results}
\scriptsize
\begin{adjustbox}{max width=1.75\textwidth,center}
\begin{tabular}{ccccccc}
\cmidrule(lr){2-7}
 & \multicolumn{6}{c}{Model Types} \\
\toprule
Noise Type & Standard & IDPA (\citeyear{wang2021tackling_instance}) & Co-teaching (\citeyear{han2018coteaching}) & CDR (\citeyear{xia2021robust}) & LS (\citeyear{wei2022smooth}) & \algsn{} (Ours) \\
\midrule
Original Data & 82.81 ± 0.79 & 83.24 ± 0.56 & \textbf{85.12 ± 0.8} & 82.49 ± 1.49 & 83.89 ± 0.91 & 83.2 ± 0.83 \\
Random Label & 65.01 ± 0.65 & 64.61 ± 0.71 & 67.18 ± 0.69 & 64.8 ± 0.91 & 66.34 ± 1.13 & \textbf{71.95 ± 2.94} \\
\bottomrule
\end{tabular}
\end{adjustbox}
\end{center}
\end{table*}
%\end{wraptable}

We also tested \algsn{} on a natural language task, in which the goal is to predict the sentiment from the review of the movie. The results are presented in Table \ref{table:imdb_results}, for which we tested two types of noise. Here, \algsn{} clearly outperforms the baselines by a significant margin, with IDPA and CDR not improving standard training. 
Additionally, we expect that when the data is not noisy, using \algsn{} should not limit performance, as we see here. 
However, it is interesting that Co-teaching performs better in the ``Original Data".
We hypothesise that the IMDB dataset contains some noisy labels (through human error), which were split uniformly across data sources; a limitation of our method that we discuss in Section \ref{sec:discussion}.

\subsection*{Imagenet results with multiple noise types}
\label{sec:imagenet_results}

We additionally present results on Imagenet, which allow us to test \algsn{} on a large-scale dataset. 
As with the experiment in Figure \ref{fig:ptbxl_aucpr_vs_corruption}, noisy sources on Imagenet contain both input and label noise. 
In Figure \ref{fig:ptbxl_aucpr_vs_corruption} all noisy sources contain two types of noise, however in the experiment presented in Table \ref{table:imagenet_results_random_label}, we allow different sources to contain different types of noise. 
These results further illustrate that the use of \algsn{} improves model performance when training data is noisy, since it achieves a greater maximum top-5 test accuracy than the standard training method.
In Appendix \ref{sec:late_training}, we also study the learning curve on this dataset to understand the effects of early-stopping used in combination with our method.

\begin{table}[ht]
\begin{center}
\caption{\textbf{Imagenet results.} Mean ± standard deviation of maximum test top-5 accuracy (\%) over $5$ repeats of standard training and \algsn{} on Imagenet data with noisy labels and noisy inputs. Here, $5$ out of $10$ sources are 100\% noisy, with three of them containing label noise and 2 containing input noise.}
\label{table:imagenet_results_random_label}
\scriptsize
\begin{tabular}{ccc}
\cmidrule(lr){2-3}
 & \multicolumn{2}{c}{Model Types} \\
\toprule
Noise Type & Standard & LAP (Ours) \\
\midrule
Input and Label Noise & 68.05 ± 0.26 & \textbf{70.61 ± 0.26} \\
\bottomrule
\end{tabular}
\end{center}
\end{table}

\subsection*{Real-world and imbalanced data sources}

\begin{table*}[ht]
\begin{center}
\caption{\textbf{Results on GoEmotions, an imbalanced sources dataset.} Mean ± standard deviation of maximum test top-5 accuracy (\%) over $5$ repeats of the methods on the GoEmotions data with noisy labels and random sentence permutation. Here, $30$ out of $82$ sources are 100\% noisy.}
\label{table:goemotions_results_random_label}
\scriptsize
\begin{adjustbox}{max width=1.75\textwidth,center}
\begin{tabular}{ccccccc}
\cmidrule(lr){2-7}
 & \multicolumn{6}{c}{Model Types} \\
\toprule
Noise Type & Standard & IDPA (\citeyear{wang2021tackling_instance})& Co-teaching (\citeyear{han2018coteaching}) & CDR (\citeyear{xia2021robust}) & LS (\citeyear{wei2022smooth}) & \algsn{} (Ours) \\
\midrule
Original Data & \textbf{80.43 ± 0.05} & 79.66 ± 0.21 & 79.7 ± 0.31 & \textbf{80.37 ± 0.1} & \textbf{80.35 ± 0.07} & \textbf{80.44 ± 0.15} \\
Random Label & 76.96 ± 0.58 & 76.05 ± 0.84 & 77.05 ± 0.7 & 76.05 ± 0.31 & 76.02 ± 0.57 & \textbf{78.74 ± 0.41} \\
\bottomrule
\end{tabular}
\end{adjustbox}
\end{center}
\end{table*}

To supplement the results of PTB-XL, we present \algsn{} on an additional dataset containing real-world data sources, GoEmotions. 
This dataset allows us to explore the scenario in which source sizes and the classes they contain vary significantly, which could be a common scenario in real-world use cases. 
As \algsn{} weights sources based on their log-likelihood, it is important to consider the robustness of our method to variations in the distribution of classes in sources, as well as their sizes. 
In the GoEmotions dataset, our training set contains source sizes in the range of $1$ to $9320$ with a mean size of $1676$ and a standard deviation of $1477$ -- providing varied source data distributions (details in Appendix \ref{sec:source_distribution_goemotions}). 
Table \ref{table:goemotions_results_random_label} demonstrates the top-5 accuracy of the methods for this task, and suggests that \algsn{} is robust to varied source data distributions in this real-world dataset. Whilst \algsn{}, CDR, Label Smoothing (LS) and standard training perform similar when trained on the original data, using \algsn{} leads to improved top-5 accuracy on the test set when random labels are introduced in training.
On GoEmotions, random labelling had small effects on the performance of the standard training method, reducing the top-5 accuracy of the test set from $80.4$ to $77.0$, however, \algsn{} was still able to significantly reduce that performance loss by achieving a top-5 accuracy of the test set of $78.7$. Here, \algsn{} produced a higher top-5 accuracy than the baselines, with less variability in the results.
In Appendix \ref{sec:straining_assumptions} we further test the imbalance in source-class distributions with an extreme example.

\subsection*{Additional results} 

Many further experiments, such as with varied numbers of sources and source sizes (\ref{sec:cifar_different_noise_results} and \ref{sec:cifar10_lots_of_sources}), model sizes (\ref{sec:cifar10n_small_network}), Imagenet (\ref{sec:tiny_imagenet_results_random_label_results} and \ref{sec:late_training}), a regression task (\ref{sec:california_housing}), the effect on late training performance (\ref{sec:late_training}) and experiments that strain the assumptions of the method (\ref{sec:straining_assumptions}) can be found in the appendix, which provides further intuitions about \algsn{} and strong evidence for its use in a wide variety of settings.

\section{Discussion}
\label{sec:discussion}

This research reveals some interesting future research directions. In our experiments, we study models of varied capacity (for example, Table \ref{table:synthetic_results} and Appendix \ref{sec:cifar10n_small_network}, and Table \ref{table:synthetic_presnet_results} and Table \ref{table:imagenet_results_random_label}), however, it is interesting to further study the effects of ill-specified models on noisy data techniques, as most methods assume that models attain smaller losses on the non-noisy data points (for example: in our work and Co-teaching \citep{han2018coteaching}), or that logits are reliable (for example: in RRL \citep{lixiong2021learning} and IDPA \citep{wang2021tackling_instance}).
In addition, we focus on the setting in which knowing the source of a data point provides additional information when learning from noisy data. However, when the noise level is uniform across \textit{all} sources, the source value does not provide additional information about a data point's likelihood of being noisy, and so in this case \algsn{} performs as well as standard training. Here, it would be beneficial to apply a noisy data technique in addition to our method, such as with RRL + LAP, studied in the experiments (Table \ref{table:synthetic_presnet_results}, Figure \ref{fig:cifar10n_acc_vs_corruption_presnet}, and Appendix \ref{sec:cifar_different_noise_results}).

The additional computation required to apply \algsn{} to a single source is $O(S + B)$ and the additional memory cost is $O(S \times H)$ (where $S$ is the number of unique sources in the dataset, $B$ is the batch size and $H$ is the length of the loss history), as we calculate the source means and standard deviations online. For multiple sources in a batch, the additional compute becomes $O(S \times S_b + B)$ (where $S_b$ is the number of unique sources in a batch). A measured time cost for a simple experiment is given in Appendix \ref{sec:compute_example}. The extra memory cost in this case is $O(S \times H + S_b \times S)$.  Although vectorised in the current implementation, $O(S \times S_b)$ could be performed with $S_b$ parallel jobs of $O(S)$, since each source calculation is independent, significantly improving its speed for larger numbers of sources. However, this is already significantly faster than some of the baselines tested, in which Co-teaching trains two models, IDPA trains a model for twice as long, and RRL applies $k$-nearest neighbours at each epoch.

Importantly, as is the case with all methods designed for training models on noisy data, consideration must be given to the underlying cause of noise. For example, in some scenarios, noise within a dataset could be attributed to observations on minority groups during data collection, rather than as a result of errors in measurements, labelling, or data transfer. This is why we began with the assumption that all data (if non-noisy) is expected to be generated from the same underlying probability distribution. If data from minority groups are not carefully considered, \textit{any} technique to learn from noisy data, or standard training could lead to an unexpected predictive bias \citep{mehrabi2021survey}. 

\section{Conclusion}

In this work, we presented \algsn{}, a method designed for training neural networks on data generated by many data sources with unknown noise levels. 
In Section \ref{sec:results}, we observed that using \algsn{} during training improves model performance when trained on data generated by a mixture of noisy and non-noisy sources, and does not hinder performance when all data sources are free of noise. 
The results also illustrated that applying the proposed method is beneficial at varied noise levels, when training on multiple sources with differing levels of noise, and whilst being robust to different types of noise. 
We also see through the results in Table \ref{table:synthetic_results}, \ref{table:synthetic_presnet_results}, \ref{table:imdb_results}, \ref{table:goemotions_results_random_label} and Figure \ref{fig:ptbxl_aucpr_vs_corruption}, and \ref{fig:cifar10n_acc_vs_corruption_presnet} that our method is applicable and robust in multiple domains with different tasks. Moreover, within the Appendix we present many further experiments, showing that: 
(1) Our method is robust to an extreme source class distribution that tests the limits of the assumptions we made when proposing \algsn{} (Appendix \ref{sec:straining_assumptions});
(2) The improvements in results translate to large-scale datasets (Imagenet), with multiple noise types (Appendix \ref{sec:imagenet_results}); 
(3) \algsn{} is robust to overfitting of noisy data, and therefore achieves significantly greater performance during late training (Appendix \ref{sec:late_training}); 
and (4) \algsn{} continues to outperform the baseline on a regression task (Appendix \ref{sec:california_housing}). 
Furthermore, our implementation (Appendix \ref{sec:software}) allows this method to be easily applied to any neural network training where data are generated by multiple sources, and our analysis in Section \ref{sec:methods} and Appendix \ref{sec:hparam_discussion} provides a detailed description of the parameters introduced and their intuitions. 

This work shows that \algsn{} provides improved model performance over the baselines in a variety of noise settings and equal performance on non-noisy data. This is achieved whilst being robust to a multitude of tasks and being cheaper to compute than the baselines designed to tackle noisy data. We therefore imagine many scenarios where \algsn{} is relevant.

\subsubsection*{Author Contributions}
\textbf{AC:} Original idea conception, software, analysis of results, writing, editing, and reviewing; 
\textbf{FP:} analysis of results, editing, and reviewing;
\textbf{TC:} editing, and reviewing;
\textbf{PB:} Supervision, analysis of results, reviewing, funding acquisition.

\subsubsection*{Acknowledgments}
This study is funded by the UK Dementia Research Institute (UKDRI) Care Research and Technology Centre funded by the Medical Research Council (MRC), Alzheimer's Research UK, Alzheimer’s Society (grant number: UKDRI–7002), and the UKRI Engineering and Physical Sciences Research Council (EPSRC) PROTECT Project (grant number: EP/W031892/1). Infrastructure support for this research was provided by the NIHR Imperial Biomedical Research Centre (BRC) and the UKRI Medical Research Council (MRC).
The funders were not involved in the study design, data collection, data analysis, or writing the manuscript.

\bibliography{lap}
\bibliographystyle{unsrtnat}

\newpage

\appendix

\section{Appendix}

\subsection{Code implementation}
\label{sec:software}

An implementation of the proposed method as well as the code to reproduce the results in this paper are made available
\begin{itemize}
    \item To reproduce the results in this paper, the code is made available here: \url{https://github.com/alexcapstick/unreliable-sources}.
    \item To implement these methods on a new task, a python package is available here: \url{https://github.com/alexcapstick/loss_adapted_plasticity}.
\end{itemize}

The experiments in this work were completed in Python 3.11, with all machine learning code written for Pytorch 2.1 \citep{paszke2017automatic}. Other requirements for running the experiments are available in the supplementary code. We make our code available under the MIT license.

All tested datasets are publicly available and easily accessible. Additionally, within the supplementary code we use the default Pytorch Dataset objects for CIFAR-10, CIFAR-100, and F-MNIST and provide Pytorch Dataset objects that will automatically download, unzip, and load the data for PTB-XL, CIFAR-10N, IMDB, and California Housing. We additionally provide code to load Tiny-Imagenet and Imagenet from a local directory since it requires an agreement before accessing. This makes reproduction of the work presented simple to perform.

The baselines were made available by the authors on GitHub: 
\begin{itemize}
    \item ARFL \citep{li2020federated}: MIT License: \url{https://github.com/lishenghui/arfl}.
    \item IDP \citep{wang2021tackling_instance}: License not provided: \url{https://github.com/QizhouWang/instance-dependent-label-noise}.
    \item Co-teaching \citep{han2018coteaching}: License not provided: \url{https://github.com/bhanML/Co-teaching}.
    \item RRL \citep{lixiong2021learning}: BSD 3-Clause License: \url{https://github.com/salesforce/RRL}.
    \item CDR \citep{xia2021robust}: License not provided: \url{https://github.com/xiaoboxia/CDR}.
    \item Label Smoothing \citep{wei2022smooth}: License not provided: \url{https://github.com/UCSC-REAL/negative-label-smoothing}.
\end{itemize} 

All experiments were carried out on a single A100 (80GB VRAM), and 32GB of RAM. The full research project required more compute than the experiments reported in the paper to design and implement \algsn{}, as well as choose training parameters for the standard training model that were then used for \algsn{} and the baselines. More information on the computation time required to run each of the main experiments is found in Appendix \ref{sec:experiment_information}, and the corresponding section of the Appendix for further experiments. In total, using this hardware, experiments took approximately 15 days to complete.

\subsection{Loss assumptions}
\label{sec:loss_assumptions}

Within this work, and for many other methods designed for tackling noisy data, we make use of the assumption that during training, neural networks learn non-noisy patterns before fitting to noisy data and therefore achieve a greater log-likelihood on non-noisy data during early training. This is discussed in length within the literature, with both theoretical and empirical justifications presented~\citep{zhang2017understanding, arpit2017closer, han2018coteaching, arazo2019unsupervised, yu2019does, shen2019learning, lixiong2021learning, wang2021tackling_instance, xia2021robust}. We also observe these training dynamics in Appendix \ref{sec:late_training} (specifically, Figure \ref{fig:imagenet_training_and_testing_curves}), in which the standard training method overfits to noisy data points significantly during later training stages compared to our method for learning from noisy data.

\subsection{Maximum tempered likelihood estimation on clean data}
\label{sec:appendix_prove_no_noise_same_as_standard}

When the dataset contains no noise and batches are large, we can show theoretically that maximising the tempered likelihood provides the same update as the unmodified maximum likelihood training at each parameter step, with a high probability.

We start with the unmodified likelihood, which is iteratively maximised. At every step, because our temperatures are calculated separately from the parameter optimisation, we can assume that they are constant for a single-parameter update. We will denote all data in the data set as $\D$, while $\D_s$ refers to all data from the source $s \in S$. Similarly, we will use the shorthand $\Pr_{\theta} ( \cdot ) = \Pr( \cdot | \theta)$.
\begin{align*}
    \arg \max_{\theta} \log \Pr_{\theta} (\D) && \text{[Unmodified likelihood]} \\
    & \hspace{-5em} = \arg \max_{\theta} \sum_{s \in S} \log \Pr_{\theta} (\D_s) & \text{[Observations are independent]} \\
    & \hspace{-5em} = \arg \max_{\theta} \sum_{s \in S} T_s \log \Pr_{\theta} (\D_s) & \text{[When } T_s = 1 ~ \forall s\in S \text{]}
\end{align*}
This provides the parameter $\theta$ that maximises the tempered likelihood. 
Therefore, if $T_s = 1$ for all steps, we maximise the same objective as in the unmodified maximum likelihood estimation at each step of the optimisation process.

Since all of the data is non-noisy, the expectation of the log-likelihood of a single observation will remain the same, including across sources. Therefore, given a sample $B$ of observations from $\D$, those observations in $B$ that are contained within each source will have the same mean log-likelihood on average. This is because the expectation of a sample mean is equal to the population mean.
\begin{equation*}
    \expt [\frac{1}{|B\cap\D_i|}\log\Pr_{\theta} (B\cap\D_i)] = \expt [\frac{1}{|B\cap\D_j|}\log\Pr_{\theta} (B\cap\D_j)] ~\hspace{1em} \forall i,j \in S
\end{equation*}
When the batch size is large, we assume that the observations in $B$ are split by source according to the proportional source sizes. Let $s_i = |\D_s|/|\D|$ be the proportional size of a source. 
If the central limit theorem is met, then in a single step, the difference of a given source's mean log-likelihood ($\mu_{i} = \frac{1}{|B\cap\D_i|}\log\Pr_{\theta} (B\cap\D_i)$) and the mean of all of the other sources' log-likelihoods ($\mu_{i'} = \frac{1}{|B\cap\D_i'|}\log\Pr_{\theta} (B\cap\D_i')$) is distributed as:
\begin{equation*}
    \mu_{i'} - \mu_{i} \sim N \left( 0,\frac{\text{Var}[\log\Pr_{\theta}(\D)]}{s_i|B|} + \frac{\text{Var}[\log\Pr_{\theta}(\D)]}{(1-s_i)|B|} \right)
\end{equation*}
By setting $\text{Var}[\log\Pr_{\theta}(\D)] = \sigma_{\text{LL}}^2$ and simplifying the fraction, this becomes:
\begin{equation*}
    \mu_{i'} - \mu_{i}  \sim N \left( 0,\frac{\sigma_{\text{LL}}^2}{s_i(1-s_i)|B|} \right)
\end{equation*}
Therefore, the probability that we increase the perceived unreliability of the source $i$ at a given step with a history length of $1$, using Equation \ref{eq:lap_method_equation} and $\lambda >0$ is:
\begin{equation}
\label{eq:prob_of_increasing_cs_no_noise}
    \Pr(\text{increase } C_s) = \Pr \left( \mu_{i'} - \mu_{i}  > \lambda \sqrt{\frac{\sigma_{\text{LL}}^2}{s_i(1-s_i)|B|}} > 0 \right) > 0.5
\end{equation}
This means that under these assumptions we expect to reduce $C_s$ at each step in the temperature calculations, and because we clip $C_s$ below at $0$, we expect $T_s = 1$ when all data is clean. When the history length is more than $1$, this can be extended by also including the expected log-likelihood over the different parameters $\theta$ at each model update step.

Equation \ref{eq:prob_of_increasing_cs_no_noise} also theoretically describes how the value of $\lambda$ influences training. This equation can be used to precisely set $\lambda$ to define the probability of incorrectly increasing $C_s$ for a given non-noisy source.

Therefore, if the batch size is large enough such that the central limit theorem applies, we expect that the maximum tempered likelihood estimation provides the same value as the maximum likelihood estimation.

\subsection{Algorithm details}
\label{sec:appendix:alg_lap_mathod}

In the following, we present the implementation of our method as gradient scaling for clarity. Given a dataset $\mathcal{D}$ of features and labels
that is generated by sources $S$, $\mathcal{S} = \{ s_1, ..., s_S \}$. We denote a subset of $\mathcal{D}$ generated by the source $s$ as $\mathcal{D}_s \subset \mathcal{D}$, 
with each data point corresponding to a single source. 

In a single update step of a model: we have the perceived \textit{unreliability} for each source $C=\{C_s\}_1^S$ (initially, all sources are considered reliable and so $ C_s = 0 ~ \forall s$); a batch of features, labels, and source values from $\D$; and a history of training losses, $L$ with length $H$ for all sources. Here, $L \in \mathbb{R}^{S \times H}$, and $l_H^s \in L$ denote the mean loss of data from source $s$ on the most recent batch containing data in $\mathcal{D}_s$ (within the subscript $H$). 
A larger $C_s$ denotes a larger estimated noise level for the source $s$. 
In each step, we update the value $C_s$ (and hence the temperature) for a source $s$ using the empirical risk of all other sources $s'$ as in Algorithm \ref{alg:lap_method}.
Further, we define $f (C_s)$ as: 

\begin{equation*}
    f(C_s) = (1-d_s) = 1-\tanh^2 (0.005 \cdot \delta \cdot  C_s )
\end{equation*}
This choice of $f$ has some nice properties discussed in Appendix \ref{sec:design_decisions}.
We refer to $d_{s}$ as the depression value; and $\delta$ is the depression strength, controlling the rate of depression.

\subsection{Further details on design decisions}
\label{sec:design_decisions}

\subsubsection*{Intuitive description of Lambda}
\label{sec:appendix_lambda_desc}

If no sources are noisy, and the distribution of negative log-likelihoods (NLLs) from the data sources forms a normal distribution, then we incorrectly reduce the temperature of a non-noisy source $s$ with probability $\Pr_\theta( \hat{L_s} \geq \lambda )$ where $L_s = - \log \Pr_{\theta} (\D_s)$ is the NLL on data from source $s$, and $\hat{L_s}$ refers to the standardisation of the NLL using the mean and variance of the NLLs of all other sources $L_{s'}$. More details in the influence of a given $\lambda$ value are given in Appendix \ref{sec:appendix_prove_no_noise_same_as_standard}.

\subsubsection*{Why use a weighted mean and standard deviation?}
Firstly, we will discuss why the weighted mean and standard deviation is used for comparing sources with each other. This is done to allow for the identification of noisy sources by our method when different data sources might have significantly different noise levels. This is the case in Figure \ref{fig:toy_example}, where we have one source producing $100\%$ noise and other sources with noise levels $5\%$ and $2.5\%$. Here, because the noise in the sources is so varied, sources with lower noise levels would likely never have an average loss more than the unweighted mean of the loss trajectory plus the threshold created from the leniency multiplied by the unweighted standard deviation (Algorithm \ref{alg:lap_method}). By weighting the mean and standard deviation with the calculated temperature, we are able to ensure that all sources with a noise level greater than $0\%$ will be discovered in a number of model update steps proportional to their noise level. This is shown in Figure \ref{fig:source_loss_with_Czeta}, where the weighted mean and standard deviation of the loss moves further to the left, as the weight reduces the influence of the unreliable source (on the right).

\subsubsection*{Why the mean + standard deviation for thresholding?} 

When calculating whether a source should be considered more or less unreliable, we need to be able to calculate the reliability compared to the other sources. We therefore use a mean and standard deviation over previous loss values, so that we can calculate a \textit{relative} reliability. As described in Section \ref{sec:methods}, it also provides an intuitive idea about the probability that a non-noisy source is incorrectly identified as noisy on a single step (if the means of loss values are assumed to follow a Gaussian distribution, which is not unreasonable according to the central limit theorem).

\subsubsection*{Why do we clip reliability at 0?}

In Algorithm \ref{alg:lap_method}, we clip $C_s$ at $0$ to ensure that when we have sources with noise levels of different magnitudes, once the sources with larger noise levels have been heavily weighted and therefore have insignificant effects on the weighted mean and standard deviation, we are able to start reducing the reliability of other noisy sources immediately, rather than waiting for $C_s$ to increase from some negative value to $0$. Also, since we use the $\tanh^2$ function to calculate the loss (or gradient) scaling, clipping $C_s$ at $0$ prevents us from introducing an error in which loss from reliable sources is scaled with the same factor as unreliable sources.

\subsection{Alternative interpretation}
\label{sec:model_interpretations}

Here, we present another interpretation of our proposed method which may provide inspiration for future research.

As discussed in Appendix \ref{sec:loss_assumptions}, the model evidence, $\Pr ( \mathcal{D} | \mathcal{M}  )$ ($\mathcal{D}$: the dataset and $\mathcal{M}$: the model) is the probability that a dataset is generated by a given model (by marginalisation of the model parameters). In our case, this could be used to calculate the probability that the data from each source $s$ is generated by the given model, that is, $\Pr ( \mathcal{D}_s | \mathcal{M}  )$ where $\mathcal{D}_s$ are the data generated by a source $s$. 
A model designed for non-noisy data should have a high $\Pr ( \mathcal{D}_s | \mathcal{M}  )$ when $\mathcal{D}_s$ has a low noise level and a low $\Pr ( \mathcal{D}_s | \mathcal{M}  )$ when $\mathcal{D}_s$ has a high noise level. 
However, $\log (\Pr ( \mathcal{D}_s | \mathcal{M}  ))$ is hard to compute for neural networks and would require an approximation. 
It can be shown that $\Pr ( \mathcal{D} | \mathcal{M}  )$ can be approximated by the \texttt{logsumexp} of the log-likelihood over a training trajectory with noisy gradients, which approximates posterior samples with SGLD or SGHMC \citep{chen2014stochastic, welling2011bayesian}. 
Within our work, when computing if a given source is noisy or non-noisy, the mean loss for this source is calculated over a training trajectory of length equal to the history length, $H$ (given by $\mu_s$ in Algorithm \ref{alg:lap_method}). 
This is then compared to a weighted mean of the loss over a training trajectory of length $H$ of all sources excluding the source under scrutiny (given by $\mu_{s'}$ in Algorithm \ref{alg:lap_method}). 
This weighted mean is related to $p(\mathcal{D}_R|\mathcal{M})$ where $\mathcal{D}_R$ represents the data from all non-noisy sources, since the weights allow us to reduce the influence of noisy sources. 
Therefore, our method is related to one in which the model evidence is calculated for each source at each step, and where the sources with a relative model evidence higher than some threshold (defined in our work through leniency, $\lambda$) have a larger influence on the model parameters when updating the weights.

\subsection{Details of the varied parameter experiments}
\label{sec:toy_example}

To build an intuition for the parameters introduced in the definition of \algsn{}, we run some synthetic examples to understand their effects on $1-d_s$ (in Equation \ref{equation:gradient_depression}). The results of this are presented in Figure \ref{fig:toy_example}. 

\parahead{Dataset.} The synthetic data is created using Scikit-Learn's make\_moons \footnote{\href{https://scikit-learn.org/stable/modules/generated/sklearn.datasets.make_moons.html}{\nolinkurl{Scikit-Learn:make\_moons}}} function, which produced $10,000$ synthetic observations with $2$ features. Each observation is assigned a label based on which ``moon" the observation corresponded to. Then, to produce synthetic sources and noise levels for each source, we randomly assign each data point a source number from $0$ to $4$, so that we have $5$ sources in total. The data points for each source were then made noisy by randomly flipping labels such that each source had a noise level of $0.0$, $0.025$, $0.05$, $0.25$, or $1.0$.

\parahead{Model and training.} A simple Multilayer Perceptron (MLP) with hidden sizes of ($100$, $100$) and ReLU activation functions is trained on this data using the Adam optimiser with a learning rate of $0.01$, a weight decay of $0.0001$, and with $(\beta_1, \beta_2) = (0.9, 0.999)$ for $50$ epochs with a batch size of $128$. Data is shuffled at each epoch before being assigned to mini-batches.

\parahead{\algsn{} parameters.} When conducting the experiments, the values of $\delta$, $H$, and $\lambda$ were varied as is described in the graphs shown in Figure \ref{fig:toy_example} to experiment with different values of these newly introduced parameters. Unless otherwise specified, the values chosen are $H=25$, $\delta=1.0$, $\lambda=1.0$.

These experiments take approximately $5$ minutes to complete when using the compute described in Appendix \ref{sec:software}.

\subsection{Further information on the datasets}
\label{sec:dataset_further_information}

The content of each of the datasets is as follows:

\begin{itemize}
    \item \textbf{CIFAR-10} and \textbf{CIFAR-100} \citep{krizhevsky2009learning} are datasets made up of \num{60000} RGB images of size $32\times32$ divided into 10 and 100 classes, with \num{6000} and $600$ images per class, respectively. The task in this dataset is to predict the correct class of a given image.
    \item \textbf{F-MNIST} \citep{xiao2017fashion} is made up of \num{70000} greyscale images of clothing of size $28\times28$ divided into \num{10} classes with \num{7000} images per class, which are flattened into observations with \num{784} features. The task of this dataset is to predict the correct class of a given image.
    \item \textbf{PTB-XL} \citep{Wagner2020} consists of \num{21837} ECG recordings of $10$ second length, sampled at $100$Hz, from \num{18885} patients, labelled by $12$ nurses. The task of this dataset is to predict whether a patient has a normal or abnormal cardiac rhythm.
    \item \textbf{CIFAR-10N} \citep{wei2022learning} is a dataset made up of all CIFAR-10 examples, but with human labelling categorised into different levels of quality. For our case, to allow for the most flexibility in experiment design, we utilised the worst of the human labels.
    \item \textbf{GoEmotions} \citep{demszky2020goemotions} is a natural language dataset in which the task is to correctly classify the emotion of Reddit comments from a possible $28$ emotions. This dataset contains \num{171820} text examples, annotated by $82$ raters, with each rater contributing somewhere between $1$ and \num{9320} labels (mean: \num{1676}, standard deviation: \num{1477}). To produce the training and test set, we randomly split the examples in the 80:20 ratio.
    \item \textbf{IMDB} \citep{maas2011imdb} is a natural language dataset with a sentiment analysis task. The goal is to correctly classify a movie review as positive or negative based on its text. It contains \num{25000} reviews in the training set and \num{25000} reviews in the testing set, divided equally between positive and negative sentiment.
    \item \textbf{MNIST} \citep{lecun1998mnist} contains \num{60000} training and \num{10000} testing images of handwritten digits in black and white with a resolution of $28 \time 28$. The goal of this dataset is to correctly identify the digit drawn in the image (from $0$ to $9$).
    \item \textbf{Tiny-Imagenet} \citep{deng2009imagenet} contains \num{110000} RGB images of size $64\times64$ of $200$ classes, with the goal of correctly classifying an image into the given class. There are \num{100000} training images and \num{10000} images in the test set.
    \item \textbf{Imagenet} \citep{deng2009imagenet} contains \num{1281167} RGB images scaled to size $64\times64$  from \num{1000} classes, with the goal of correctly classifying an image into the given class. There are \num{1231167} training images and \num{50000} images in the test set.
    \item \textbf{California Housing} \citep{pace1997sparse} contains \num{20640} samples of house values in California districts, with the aim of predicting the median house value from the US census data for that region. We randomly split the data into training and testing with an $8:2$ ratio on each repeat.
\end{itemize}

\parahead{Dataset licenses.} CIFAR-10, CIFAR-100, and F-MNIST are managed under the MIT License. PTB-XL is made available with the Creative Commons Attribution 4.0 International Public License~\footnote{\url{https://creativecommons.org/licenses/by/4.0/}.}, CIFAR-10N is available with the Creative Commons Attribution-Non-Commercial 4.0 International Public License~\footnote{\url{https://creativecommons.org/licenses/by-nc/4.0/}.}, GoEmotions is made available under the Apache-2.0 license~\footnote{\url{https://github.com/google-research/google-research}.}, and MNIST is available under the the Creative Commons Attribution-Share-Alike 3.0 license~\footnote{\url{https://creativecommons.org/licenses/by-sa/3.0/}}.
IMDB\footnote{\url{https://ai.stanford.edu/~amaas/data/sentiment/}} and California Housing\footnote{\url{https://www.dcc.fc.up.pt/~ltorgo/Regression/cal_housing.html}} are made publicly available, and Tiny-Imagenet and Imagenet are available after agreeing to the access terms \footnote{\url{https://www.image-net.org/download}}.

\parahead{Dataset availibility.} As mentioned in Appendix \ref{sec:software}, all datasets tested are publicly available and easily accessible. Additionally, within the supplementary code we use the default Pytorch Dataset objects for CIFAR-10, CIFAR-100, F-MNIST, and MNIST, and provide Pytorch Dataset objects that will automatically download, unzip, and load the data for PTB-XL, CIFAR-10N, GoEmotions, IMDB, and California Housing. We additionally provide code to load Tiny-Imagenet and Imagenet from a local directory since it requires an agreement before accessing. This makes reproduction of the work presented easy to perform.

\subsection{Experiments in detail}
\label{sec:experiment_information}
For each of the experiments presented in Section \ref{sec:results}, we will now describe the dataset, models, and training in detail. 

\subsubsection*{Model architectures}

Within our experiments, we tested many model architectures, described below. Due to limitations in computation and to allow for our work to be easily reproducible, we use different levels of model capacity, which additionally illustrates that our method is applicable in varied settings. All models are also available in the supplementary code, implemented in Pytorch.

\begin{itemize}

    \item \textbf{Low capacity multilayer perceptron (MLP):} The MLP used in Section \ref{sec:results} consists of 3 linear layers that map the input to dimension sizes of $16$, $16$, and the number of classes. In between these linear layers we apply dropout with a probability of $0.2$, and a ReLU activation function. For our experiments on California Housing, we used hidden sizes of $32$, $32$, $32$, and $1$ (for the output value) with ReLU activation functions.
    
    \item \textbf{Low capacity CNN:} This model consisted of $3$ convolutional blocks followed by $2$ fully connected layers. Each convolutional block contained a convolutional layer with kernel size of $3$ with no padding, and a stride and dilation of $1$; a ReLU activation function; and a max pooling operation with kernel size of $2$. The linear layers following these convolutional blocks maps the output to a feature size of $64$ and then the number of classes, with a ReLU activation function in between.

    \item \textbf{High capacity CNN with contrastive learning:} This CNN is inspired by the model used for the CIFAR-10 experiments in \citet{lixiong2021learning} and requires considerably more compute. It is based on the ResNet architecture \citep{he2016deep}, except that it uses a pre-activation version of the original ResNet block. This block consists of a batch normalisation operation, a convolutional layer (of kernel size $3$, a padding of $1$, and varied stride, without a bias term), another batch normalisation operation, and another convolutional layer (with the same attributes). Every alternate block contains a skip connection (after a convolutional layer, with a kernel size of $1$ and a stride of $2$, had been applied to the input). The input passes through a convolutional layer before the blocks. This is followed by a linear layer that maps the output of the convolutions to the number of classes. This model also contains a data reconstruction component that allows for a unsupervised training component.

    \item \textbf{ResNet 1D:} This model is designed for time-series classification and is based on the 2D ResNet model \citep{he2016deep}, except with 1D convolutions and pooling. The model is made up of $4$ blocks containing two convolutional layers split by a batch normalisation operation, ReLU activation function, and a dropout layer. These convolutional blocks reduce the resolution of their input by a quarter and increase the number of channels linearly by the number of input channels in the first layer of the model. Each block contains a skip connection that is added to the output of the block before passing through a batch normalisation operation, ReLU activation function and dropout layer. This is followed by a linear layer that transforms the output from the convolutional blocks to the number of output classes.

    \item \textbf{ResNet 2D:} This model is designed for image classification and is exactly the ResNet 20 architecture presented in \citet{he2016deep} or the ResNet 18 or Resnet 50 implementation in available with Pytorch~\footnote{\href{https://pytorch.org/vision/main/models/resnet.html}{\nolinkurl{Pytorch:ResNet}}}.

    \item \textbf{Transformer Encoder:} \citep{vaswani2017attention} This model is designed for the natural language based emotion prediction task given by the GoEmotions dataset. This model uses an embedding layer of size \num{256}, positional encoding, and \num{2} transformer encoder layers (based on the implementation in Pytorch~\footnote{\href{https://pytorch.org/docs/stable/generated/torch.nn.TransformerEncoderLayer.html}{\nolinkurl{Pytorch:TransformerEncoderLayer}}}) with \num{4} heads, and an embedding size of $256$. This is followed by a linear layer that maps the output from the transformer encoder to the $28$ emotion classes.

    \item \textbf{LSTM:} \citep{hochreiter1997long} The natural language model contains an embedding layer, which maps tokens to vectors of size \num{256}, a predefined LSTM module from Pytorch~\footnote{\href{https://pytorch.org/docs/stable/generated/torch.nn.LSTM.html}{\nolinkurl{Pytorch:LSTM}}} with $2$ layers and a hidden size of $512$, a dropout layer with a probability of $0.25$, and a fully connected layer mapping the output from the LSTM module to $2$ classes.
    
\end{itemize}

\subsubsection*{Datasets and model training}

\parahead{Results in Table \ref{table:synthetic_results}.} To produce the results in Table \ref{table:synthetic_results}, we used two different models applied to three different datasets, with 7 noise settings.

\begin{itemize}
    \item \textbf{CIFAR-10 and CIFAR-100:} All training data is randomly split into $10$ sources, with $4$ and $2$ sources chosen to be noisy for CIFAR-10 and CIFAR-100, respectively. In this experiment, these noisy sources are chosen to be $100\%$ noisy so that we can understand how our method performs on data containing highly noisy sources. Noise is introduced based on the description given in Section \ref{sec:methods}. The ResNet 20 model described above is trained on this data and tasked with predicting the image class using cross-entropy loss. The model is trained for $40$ epochs in both cases, with the SGD optimiser and a learning rate of $0.1$, a momentum of $0.9$, and a weight decay of $0.0001$, in batches of size $128$. When training with \algsn{}, we use $H=50$, $\delta=1.0$, and $\lambda=1.0$ for CIFAR-10 and $H=25$, $\delta=0.5$, and $\lambda=1.0$ for CIFAR-100 chosen using the validation data (which are made noisy using the same procedure as the training data) after training on data with random label noise. When training on CIFAR-100 we found that using a warm-up of $100$ steps improved performance.
    \item \textbf{F-MNIST:} 
    All training data is randomly split into $10$ sources, with $6$ chosen to be $100\%$ noisy.
    The low capacity MLP model is trained for $40$ epochs using the Adam optimiser, and a learning rate of $0.001$ on batches of size $200$. When training with \algsn{}, we use $H=25$, $\delta=0.5$, and $\lambda=0.5$ chosen using the validation data (which is made noisy using the same procedure as the training data) after training on data with random label noise.
\end{itemize}

In our implementation, each data point contained an observation, the generating data source, and a target.  
The features and labels were passed to the model for training, and the sources were used to calculate the weighting to apply to the loss at each data point.

To allow fairness in comparison with ARFL, global updates on this model were performed the same number of times as the number of epochs other models were trained for, with all clients being trained on data for a single epoch before each global update. All other parameters for the ARFL model are kept the same. 
When IDPA was tested, the parameters were chosen as in \citet{wang2021tackling_instance} using the default parameters in the implementation.
When testing ``Co-teaching", the parameters are chosen as in \citet{han2018coteaching} where possible, and scaled proportionally by the change in the number of epochs between their setting and our setting where they depended on the total number of epochs. The forgetting rate was set as the default given in the implementation code, $0.2$, since we do not assume access to the true noise rate.
For CDR, the parameters for CIFAR-10 and 100 were chosen as given in \citet{xia2021robust}, and for F-MNIST, we used the same parameters as CIFAR-10.
When testing Label Smoothing~\citep{wei2022smooth}, we used a smoothing rate of $0.1$ for CIFAR-10, CIFAR-100, and F-MNIST as we found that using negative smoothing had adverse effects on accuracy.

Note that when training ARFL, since it is a federated learning method, each client is trained on sources separately.

These experiments take approximately $2$ days to complete when using the compute described in Appendix \ref{sec:software}.

For all of the following experiments, to reduce the computational cost, we fixed the parameters of \algsn{} to $H=25$, $\delta=1.0$, and $\lambda=0.8$. In practice, optimising these parameters should allow improved performance when using \algsn{} over those presented in this work.

\parahead{Results in Table \ref{table:synthetic_presnet_results}.} To produce the results presented here, all training data is randomly divided into $10$ sources, with $4$ chosen to be $100\%$ noisy. The high capacity CNN with contrastive learning presented in \citet{lixiong2021learning} is trained for $25$ epochs, with all other parameters kept as in the original work. In addition to this model, a version (with the same parameters) is trained using \algsn{} with $H=25$, $\delta=1.0$, and $\lambda=0.8$. These models were trained on batch sizes of $128$ using stochastic gradient descent with a learning rate of $0.02$, momentum of $0.9$, and weight decay $0.0005$. Data is randomly assigned to mini-batches, with each batch containing multiple sources.

These experiments take approximately $3$ days to complete when using the compute described in Appendix \ref{sec:software}.

\parahead{Results in Figure \ref{fig:ptbxl_aucpr_vs_corruption}.} This experiment allowed us to test the performance of using \algsn{} on different numbers of sources with varied noise levels. Firstly, data is divided into $12$ sources based on the clinician performing the labelling of the ECG recording. Then, for a given number of sources (increasing along the $x$ axis), the noisy sources have noise levels that are set at linear intervals between $25\%$ and $100\%$. These sources are made noisy following the suggestions in \citet{emiwong2012} to simulate electromagnetic interference and using label flipping to simulate human error in labelling. Based on this data, we train the ResNet 1D model for $40$ epochs using the Adam optimiser with a learning rate of $0.001$ and a batch size of $64$. This model is trained with and without \algsn{} with $H=25$, $\delta=1.0$, and $\lambda=0.8$. Data is randomly assigned to mini-batches, with each batch containing multiple sources.

These experiments take approximately $16$ hours to complete when using the compute described in Appendix \ref{sec:software}.

\parahead{Results in Figure \ref{fig:cifar10n_acc_vs_corruption_presnet}.} Here, CIFAR-10 is used as the features and the labelling collected in \citet{wei2022learning} are used as the noisy labels to produce CIFAR-10N. 
The data is first randomly split into $10$ sources. Then, for a given number of sources (increasing along the $x$ axis), the noisy sources have noise levels that are set at linear intervals between $25\%$ and $100\%$, replacing the true CIFAR-10 labels with the real noisy labels from \citet{wei2022learning}. The high capacity CNN with contrastive learning presented in \citet{lixiong2021learning} is trained for $25$ epochs, with all other parameters kept as in the original work. As before, a version of this model (with the same parameters) is trained using \algsn{} with $H=25$, $\delta=1.0$, and $\lambda=0.8$. These models are trained on batch sizes of $128$ using stochastic gradient descent with a learning rate of $0.02$, momentum of $0.9$, and weight decay $0.0005$. Data is randomly assigned to mini-batches, with each batch containing multiple sources.

These experiments take approximately $30$ hours to complete when using the compute described in Appendix \ref{sec:software}.

\parahead{Results in Table \ref{table:imdb_results}.} In this set of experiments, we want to test the use of \algsn{} in a natural language setting. Firstly, we load the training and testing sets and randomly split the training set into $10$ sources uniformly. We then choose $4$ of the sources to be $100\%$ unreliable and introduce noise through random labelling and random permuting the order of the text. We truncate or extend each review such that it contains exactly $256$ tokens. We use an LSTM to predict the sentiment of movie reviews by training the model for $40$ epochs with a batch size of $128$, a leaning rate of $0.001$, and the Adam optimiser \citep{Kingma2014}. A version of this model (with the same parameters) is also trained using \algsn{} with $H=25$, $\delta=1.0$, and $\lambda=0.8$. As before, data is randomly assigned to mini-batches, with each batch containing multiple sources. 
Since IDPA, Co-teaching, CDR, and Label Smoothing are not bench-marked on this dataset, for the baseline-specific parameters, we used the same values as given for CIFAR-10.

These experiments take approximately $12$ hours to complete when using the compute described in Appendix \ref{sec:software}.

\parahead{Results in Table \ref{table:goemotions_results_random_label}.} This experiment is designed to test the performance of \algsn{} on a dataset with real-world data sources with a significant imbalance in both their size and class distributions (for further details in Appendix \ref{sec:source_distribution_goemotions}) to better understand the robustness of our proposed method, since it calculates the log-likelihood of sources during training. We chose \num{30} of the total \num{82} raters to produce noisy labels. The Transformer Encoder was trained for \num{25} epochs with a batch size of \num{256} and a learning rate of \num{0.001} using the Adam optimiser \citep{Kingma2014}.  A version of this model (with the same parameters) is also trained using \algsn{} with $H=25$, $\delta=1.0$, and $\lambda=0.8$. As before, data is randomly assigned to mini-batches, with each batch containing multiple sources. 
Since IDPA, Co-teaching, CDR, and Label Smoothing are not bench-marked on this dataset, for the baseline-specific parameters, we used the same values as given for CIFAR-10.

These experiments take approximately $12$ hours to complete when using the compute described in Appendix \ref{sec:software}.

\parahead{Results in Table \ref{table:imagenet_results_random_label}.} For this experiment, we train a ResNet 50 architecture using the baseline training method provided for Imagenet in Pytorch\footref{footnote:imagenet_training_script}.
This trains the model for $90$ epochs with a batch size of $256$ using stochastic gradient descent with an initial learning rate of $0.1$, momentum of $0.9$, and weight decay of $0.0001$, as well as a learning rate scheduler that multiplies the learning rate by $0.1$ every $30$ epochs.
We divide the Imagenet data into $10$ sources, of which $5$ are chosen to be noisy ($2$ containing input noise and $3$ containing label noise). The images are loaded as 8-bit integer arrays and interpolated to $(64,64)$ before the input noise is synthesised by randomly adding uniform integers from [-64, 64]. They are then transformed to 32-bit floats and normalised using the mean and standard deviation available on the same Pytorch training script. Label noise is added by randomly replacing labels in noisy sources. This dataset is then split into the ratio $9:1$ to produce a validation set for the analysis in Appendix \ref{sec:late_training}.
A version of the ResNet 50 model (with the same parameters) is also trained using \algsn{} with $H=25$, $\delta=1.0$ and $\lambda=0.8$. As before, data is randomly assigned to mini-batches, with each batch containing multiple sources. These experiments are repeated $5$ times for each setting.

These experiments take approximately $32$ hours to complete when using the compute described in Appendix \ref{sec:software}.

\subsection{Assessing the sensitivity of \algsn{} to the hyperparameters}
\label{sec:hparam_discussion}

\begin{figure*}[ht]
     \centering
     \begin{subfigure}[b]{0.32\textwidth}
         \centering
         \includegraphics[width=\textwidth]{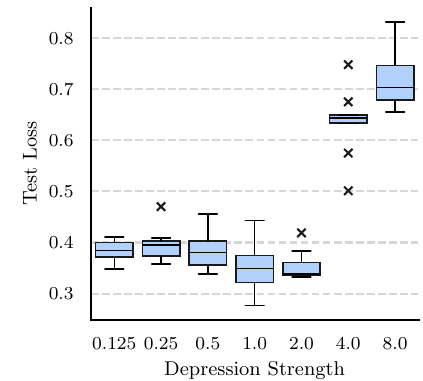}
         \caption{\textbf{Depression Strength, $\delta$.}}\label{fig:toy_example_hparam_depression_strength_boxplot}
     \end{subfigure}
     \begin{subfigure}[b]{0.32\textwidth}
         \centering
         \includegraphics[width=\textwidth]{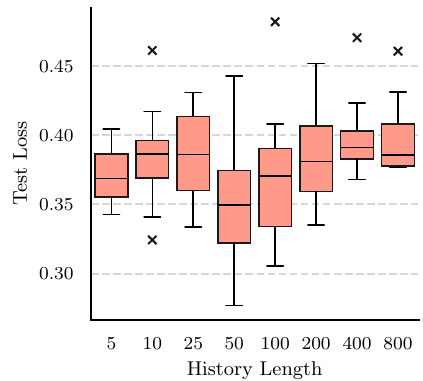}
         \caption{\textbf{History Length, $H$.}}
         \label{fig:toy_example_hparam_history_length_boxplot}
     \end{subfigure}
     \begin{subfigure}[b]{0.32\textwidth}
         \centering
         \includegraphics[width=\textwidth]{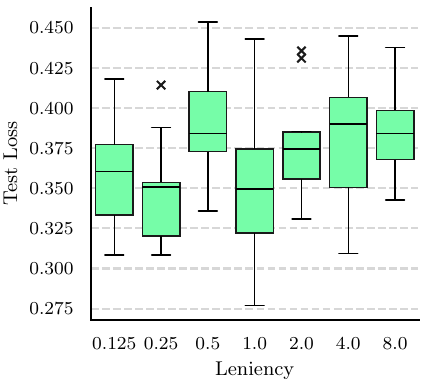}
         \caption{\textbf{Leniency, $\lambda$.}}
         \label{fig:toy_example_hparam_leniency_boxplot}
     \end{subfigure}
    \caption{\textbf{Sensitivity to Hyperparameters.} Here, we demonstrate the senstivitiy of the model performance based on the \algsn{} hyperparamters introduced in Section \ref{sec:methods} by varying their values whilst keeping the others fixed at $(\delta, H, \lambda) = (1.0, 50, 1.0)$.}
    \label{fig:senstivity_to_hyperparameters}
\end{figure*}

In Figure \ref{fig:senstivity_to_hyperparameters} we present the change in cross entropy loss on the test set of the synthetic data used in Figure \ref{fig:toy_example} and described in Section \ref{sec:toy_example}. In this experiment the depression strength $\delta$, history length $H$, and leniency $\lambda$ were kept at $(\delta, H, \lambda) = (1.0, 50, 1.0)$ unless otherwise specified. 

\parahead{Depression strength.} We observe that significantly increasing the depression strength $\delta$ can lead to a noticeable increase in test loss, particularly when $\delta \geq 2.0$ (Figure \ref{fig:toy_example_hparam_depression_strength_boxplot}). This is because large values of $\delta$ force training contributions from noisy data sources to be reduced early in training (as shown in Figure \ref{fig:toy_example}), whilst they might still be useful for learning a robust model. Suppressing them too quickly might prevent the model from learning important patterns in the noisy data which are useful for predicting on the test set, leading to worse performance. This highlights the need to choose a depression strength $\delta$ that strikes the right balance between filtering noise and learning from all available data. We find that a value of $\delta=1.0$ generally performs well.

\parahead{History length.} When studying the history length $H$, we see that the test loss remains relatively stable across the range of values (Figure \ref{fig:toy_example_hparam_history_length_boxplot}). There is some improvement for the history length values of $H=50$ and $H=100$, but there is little difference in performance between the use of large and small values of $H$. This indicates that whilst increasing $H$ allows for \algsn{} to consider more contextual information when calculating the source weighting, it also makes our method slower to react to rapid changes in the training loss due to the significantly larger history being considered. We find that a $H$ value of $25$ to $50$ generally performs well.

\parahead{Leniency.} In Figure \ref{fig:toy_example_hparam_leniency_boxplot}, we present the test loss as we vary the value of leniency $\lambda$. 
The loss values for this parameter remain relatively stable across its different values, suggesting that the model is fairly robust to variations in leniency. 
It may not be a critical parameter for tuning in this particular set up, since Figure \ref{fig:toy_example} shows that for all leniency values tested the two noisiest sources were heavily weighted during training. 
In scenarios where noise levels are very low, it might be necessary to reduce the leniency to capture them. 
In contrast, when it is believed that patterns can be learnt from noisy data, it might be beneficial to use a larger leniency value, which intuitively lengthens the amount of time before noisy sources are weighted (Figure \ref{fig:toy_example}).
In general, we find that a value of $\lambda = 1.0$ performs well.

These experiments take approximately $5$ minutes to complete when using the compute described in Appendix \ref{sec:software}.

\subsection{Table \ref{table:synthetic_results} with percentage difference in values}
\label{sec:synthetic_results_percentage_difference}

\begin{table*}[ht]
\setlength{\tabcolsep}{3pt}
\begin{center}
\caption{\textbf{Comparison of \algsn{} with the baselines.} Mean ± standard deviation of the percentage difference of the maximum test accuracy (\%) of $5$ repeats of the baselines and \algsn{} on synthetic data with different noisy types. For CIFAR-100 these numbers represent the top 5 accuracy. For CIFAR-10, CIFAR-100, and F-MNIST, the number of noisy sources are $4$, $2$, and $6$ out of $10$ respectively. Unreliable sources are each 100\% noisy. All values in bold are within 1 standard deviation of the maximum score.}
\label{table:synthetic_results_percentage_difference}
\scriptsize
\begin{adjustbox}{max width=1.75\textwidth,center}
\begin{tabular}{ccccccccc}
\cmidrule(lr){3-9}
 & & \multicolumn{7}{c}{Model Types} \\
\toprule
 & Noise Type & Standard & ARFL (\citeyear{li2022auto}) & IDPA (\citeyear{wang2021tackling_instance})& Co-teaching (\citeyear{han2018coteaching}) & CDR (\citeyear{xia2021robust}) & LS (\citeyear{wei2022smooth}) & \algsn{} (Ours) \\
\midrule
\multirow{7}{*}{\rot{CIFAR-10}} & Original Data & - & -3.76\% ± 2.40 & \textbf{2.66\% ± 1.53} & \textbf{2.86\% ± 0.45} & -0.79\% ± 2.03 & 0.84\% ± 1.23 & -0.38\% ± 1.53 \\
 & Chunk Shuffle & - & -5.12\% ± 1.59 & 0.77\% ± 0.49 & \textbf{3.68\% ± 1.50} & -2.89\% ± 3.05 & -2.79\% ± 1.84 & \textbf{3.02\% ± 2.22} \\
 & Random Label & - & 3.10\% ± 2.45 & -17.90\% ± 3.40 & 1.15\% ± 1.43 & -3.22\% ± 3.44 & 0.15\% ± 4.01 & \textbf{8.34\% ± 3.93} \\
 & Batch Label Shuffle & - & -2.31\% ± 3.95 & -0.80\% ± 1.68 & 4.29\% ± 2.08 & -0.76\% ± 2.18 & -2.00\% ± 2.74 & \textbf{7.59\% ± 2.57} \\
 & Batch Label Flip & - & -1.40\% ± 3.67 & 3.77\% ± 2.10 & 6.06\% ± 1.05 & -0.77\% ± 3.30 & 1.90\% ± 3.59 & \textbf{10.04\% ± 0.85} \\
 & Added Noise & - & -2.33\% ± 1.51 & \textbf{1.02\% ± 2.23} & \textbf{2.94\% ± 0.91} & -2.72\% ± 2.79 & -1.05\% ± 2.28 & \textbf{3.43\% ± 2.41} \\
 & Replace With Noise & - & -7.92\% ± 2.26 & -1.12\% ± 2.70 & \textbf{1.50\% ± 0.42} & -2.42\% ± 1.50 & 0.26\% ± 1.52 & \textbf{1.46\% ± 1.63} \\
 \midrule
\multirow{7}{*}{\rot{CIFAR-100}} & Original Data & - & -20.42\% ± 3.05 & \textbf{2.42\% ± 0.77} & 0.41\% ± 1.84 & -0.02\% ± 2.36 & -0.67\% ± 1.91 & 0.10\% ± 1.91 \\
 & Chunk Shuffle & - & -17.59\% ± 3.31 & -2.95\% ± 1.83 & \textbf{1.06\% ± 1.15} & -0.29\% ± 1.84 & -1.03\% ± 1.33 & \textbf{2.26\% ± 1.72} \\
 & Random Label & - & -16.20\% ± 6.51 & -15.13\% ± 1.72 & 4.93\% ± 1.96 & -2.06\% ± 1.89 & -0.07\% ± 0.90 & \textbf{18.40\% ± 2.38} \\
 & Batch Label Shuffle & - & -10.15\% ± 7.25 & -0.68\% ± 3.85 & \textbf{7.03\% ± 1.49} & 1.38\% ± 1.39 & -0.08\% ± 2.65 & \textbf{6.96\% ± 0.93} \\
 & Batch Label Flip & - & -8.64\% ± 6.29 & 4.41\% ± 3.94 & \textbf{11.57\% ± 2.15} & -3.03\% ± 2.58 & -2.00\% ± 3.63 & \textbf{12.84\% ± 2.60} \\
 & Added Noise & - & -10.60\% ± 6.71 & -1.27\% ± 2.35 & 1.14\% ± 1.34 & -0.06\% ± 2.33 & 0.15\% ± 1.18 & \textbf{3.70\% ± 2.24} \\
 & Replace With Noise & \textbf{-} & -13.02\% ± 7.46 & -2.48\% ± 2.26 & \textbf{1.12\% ± 0.75} & \textbf{1.11\% ± 3.49} & \textbf{-0.23\% ± 2.76} & \textbf{1.46\% ± 3.72} \\
 \midrule
\multirow{7}{*}{\rot{F-MNIST}} & Original Data & \textbf{-} & -1.95\% ± 0.83 & \textbf{0.07\% ± 0.71} & -5.44\% ± 1.34 & -1.44\% ± 2.87 & \textbf{0.13\% ± 0.67} & \textbf{-0.13\% ± 0.31} \\
 & Chunk Shuffle & - & 0.78\% ± 2.35 & 0.47\% ± 3.93 & -3.06\% ± 1.88 & -0.81\% ± 4.05 & 1.39\% ± 2.17 & \textbf{6.61\% ± 3.46} \\
 & Random Label & \textbf{-} & \textbf{2.28\% ± 11.45} & \textbf{0.75\% ± 6.82} & \textbf{2.79\% ± 15.20} & \textbf{0.30\% ± 18.72} & \textbf{-0.85\% ± 13.32} & \textbf{8.21\% ± 10.42} \\
 & Batch Label Shuffle & \textbf{-} & -4.29\% ± 2.05 & \textbf{0.20\% ± 0.57} & -0.42\% ± 0.29 & \textbf{0.28\% ± 0.53} & \textbf{-0.02\% ± 0.47} & \textbf{-0.24\% ± 0.43} \\
 & Batch Label Flip & - & -1.67\% ± 4.39 & \textbf{2.61\% ± 2.31} & 0.26\% ± 2.64 & \textbf{1.33\% ± 2.29} & \textbf{0.43\% ± 2.28} & -0.43\% ± 2.86 \\
 & Added Noise & \textbf{-} & -5.86\% ± 3.02 & \textbf{0.77\% ± 4.43} & -2.71\% ± 3.78 & -2.09\% ± 4.25 & \textbf{1.53\% ± 3.13} & \textbf{1.14\% ± 3.75} \\
 & Replace With Noise & - & -7.07\% ± 1.66 & 1.10\% ± 1.08 & -1.26\% ± 1.20 & -1.91\% ± 1.20 & 0.22\% ± 0.57 & \textbf{3.99\% ± 1.05} \\
\bottomrule
\end{tabular}
\end{adjustbox}
\end{center}
\end{table*}

Table \ref{table:synthetic_results_percentage_difference} shows the values presented in Table \ref{table:synthetic_results} as a percentage difference from the standard training method. This more clearly demonstrates the size of the accuracy improvement from using \algsn{} on CIFAR-10, CIFAR-100, and F-MNIST.

In particular, using \algsn{} leads to substantial improvements over the baseline in more challenging scenarios, such as with random labelling noise or batch label flipping. For example, on CIFAR-100 with random label noise, \algsn{} achieves a $21.02\%$ improvement in top-5 accuracy over standard training, significantly outperforming baselines such as IDPA. Even in cases where \algsn{} is not the highest performing method, its performance remains comparable and often within a few percentage points of the best results.

This illustrates \algsn{}'s robustness across varied datasets and noise types, showing that it consistently maintains or improves accuracy in high-noise conditions where other models struggle.

\subsection{RRL + \algsn{} with varied noise.}
\label{sec:cifar_different_noise_results}

\begin{table*}[ht]
\begin{center}
\caption{\textbf{Different noise level and number of sources.} Mean ± standard deviation (\%) of the \textit{percentage difference} in maximum test accuracy between RRL + \algsn{} and RRL (\citeyear{lixiong2021learning}) over $5$ repeats when training a model on CIFAR-10 for different noise levels and numbers of sources. Here a positive value represents an improvement in accuracy when using \algsn{}. Here, $U$ corresponds to the number of unreliable sources out of 10 sources in total. Batch label flipping was used to introduce noise.}
\label{table:cifar_different_noise_results}
\scriptsize
\begin{tabular}{ccccc}
\cmidrule(lr){2-5}
 & \multicolumn{4}{c}{Noise Level} \\
\toprule
$U$ & 25\% & 50\% & 75\% & 100\% \\
\midrule
2 & 0.88\% ± 0.19 & 1.61\% ± 0.60 & 2.95\% ± 0.64 & 3.16\% ± 0.51 \\
4 & 0.09\% ± 0.71 & 1.25\% ± 0.58 & 2.01\% ± 0.96 & 5.24\% ± 1.24 \\
6 & 0.02\% ± 0.43 & 0.46\% ± 0.50 & 3.74\% ± 0.80 & 11.61\% ± 1.66 \\
\bottomrule
\end{tabular}
\end{center}
\end{table*}

Table \ref{table:cifar_different_noise_results} shows the percentage difference in accuracy when using \algsn{} over not using \algsn{} for different noise levels and number of noisy sources. 

In these experiments, the data in CIFAR-10 is randomly divided into $10$ sources, the row of Table \ref{table:cifar_different_noise_results} defining the number of noisy sources, of which all have a noise rate as given by the column of Table \ref{table:cifar_different_noise_results}. As before, the high capacity CNN with contrastive learning presented in \citet{lixiong2021learning} is trained for $25$ epochs, with all other parameters kept as in the original work. In addition, a version of this model (with the same parameters) is trained using \algsn{} with $H=25$, $\delta=1.0$, and $\lambda=0.8$. These models are trained on batch sizes of $128$ using stochastic gradient descent with a learning rate of $0.02$, momentum of $0.9$, and weight decay $0.0005$. Data is randomly assigned to mini-batches, with each batch containing multiple sources.

When the whole dataset noise level is small (i.e., 2 sources with 25\% noise), there are small performance increases when using \algsn{} -- probably because here noisy sources have small impacts on the performance of models trained without \algsn{}. Furthermore, for lower noise levels, increasing the number of noisy sources reduced the performance improvement. This is likely because \algsn{} is reducing noisy source training contributions early, when there is still information to learn. This can be remedied by increasing the leniency ($\lambda$); however, since we were limited by the computation, the parameters of \algsn{} were fixed across the experiments.

These experiments take approximately $5$ days to complete when using the compute described in Appendix \ref{sec:software}.

\subsection{CIFAR-10N with a smaller neural network}
\label{sec:cifar10n_small_network}

\begin{figure}[ht]
    \centering
    \includegraphics[width=\linewidth]{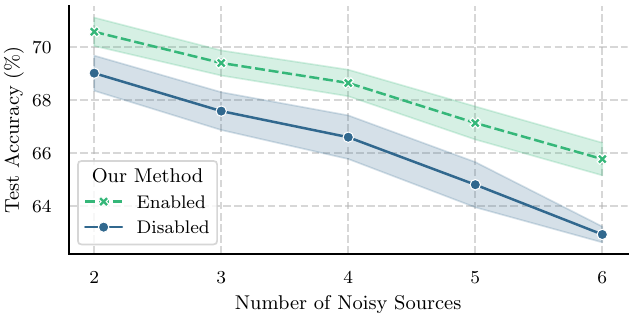}
    \caption{\textbf{Accuracy values on CIFAR-10N for an increasing number of noisy sources with a lower capacity model.} The lines and error bands represent the mean and standard deviation of the maximum test accuracy for each of the 5 repeats with random allocation of noisy sources. The noise of the sources increases linearly from $25\%$ to $100\%$ for each number of noisy sources. In total, there are $10$ sources. Here we test the low capacity CNN (Appendix \ref{sec:experiment_information}).}
    \label{fig:cifar10n_acc_vs_corruption}
\end{figure}

In Figure \ref{fig:cifar10n_acc_vs_corruption}, we show the results of the experiment presented in Figure \ref{fig:cifar10n_acc_vs_corruption_presnet} except when using the low capacity CNN described in Appendix \ref{sec:experiment_information}. 
The results of this experiment follow the same trend as in Figure \ref{fig:cifar10n_acc_vs_corruption_presnet}, which assures us of the applicability of \algsn{} in a variety of settings, where models with lower capacity are used.

These experiments take approximately $4$ hours to complete when using the compute described in Appendix \ref{sec:software}.

\subsection{CIFAR-10 with a with large numbers of sources}
\label{sec:cifar10_lots_of_sources}

\begin{table*}[ht]
\setlength{\tabcolsep}{3pt}
\begin{center}
\caption{\textbf{Comparison of \algsn{} with the baselines with varied numbers of sources.} Mean ± standard deviation of maximum test accuracy (\%) of $5$ repeats of the baselines and \algsn{} on synthetic data with different noisy types and numbers of sources. The number of noisy sources ranges from $10$ to \num{50000}, corresponding to ``\algsn{}-$n$", where $n$ denotes the number of sources in the training data. For all experiments, $40\%$ of the sources were chosen as unreliable, with 100\% noise rate. Since there are \num{50000} training data points, \algsn{}-\num{50000} corresponds to \algsn{} acting over sources that have size $1$: i.e the standard noisy data setting.}
\label{table:synthetic_results_lots_of_sources}
\scriptsize
\begin{adjustbox}{max width=1.75\textwidth,center}
\begin{tabular}{ccccccccc}
\cmidrule(lr){2-9}
 & \multicolumn{8}{c}{Model Types} \\
\toprule
 Noise Type & Standard & IDPA (\citeyear{wang2021tackling_instance}) & Co-teaching (\citeyear{han2018coteaching}) & CDR (\citeyear{xia2021robust}) & LS (\citeyear{wei2022smooth}) & \algsn{}-10 & \algsn{}-\num{1250} & \algsn{}-\num{50000} \\
\midrule
Original Data & 69.46 ± 1.12 & \textbf{70.86 ± 0.32} & 67.73 ± 0.83 & \textbf{70.88 ± 0.51} & 70.2 ± 0.66 & 69.6 ± 1.08 & 69.57 ± 1.07 & 70.28 ± 0.5 \\
Chunk Shuffle & 65.92 ± 1.32 & 66.34 ± 0.99 & 63.55 ± 0.41 & 66.18 ± 0.81 & 66.22 ± 0.84 & \textbf{66.92 ± 0.56} & 66.08 ± 1.02 & 65.47 ± 0.98 \\
Random Label & 60.85 ± 1.67 & 57.93 ± 1.59 & 61.37 ± 1.01 & 61.23 ± 1.16 & 61.1 ± 1.25 & \textbf{66.5 ± 0.43} & \textbf{66.15 ± 1.16} & \textbf{66.49 ± 0.65} \\
Batch Label Shuffle & 62.23 ± 0.62 & 60.08 ± 0.13 & 63.9 ± 0.44 & 64.63 ± 0.88 & 64.36 ± 1.02 & 66.09 ± 0.65 & \textbf{66.71 ± 0.48} & 65.73 ± 1.13 \\
Batch Label Flip & 59.94 ± 2.48 & 61.02 ± 1.9 & 63.05 ± 2.43 & 61.57 ± 1.06 & 61.84 ± 0.92 & 64.48 ± 1.11 & \textbf{66.42 ± 1.22} & \textbf{65.41 ± 1.65} \\
Added Noise & 59.86 ± 1.38 & 61.19 ± 0.33 & 59.59 ± 1.37 & 60.38 ± 0.84 & 60.05 ± 1.05 & \textbf{63.58 ± 1.03} & 59.52 ± 1.06 & 60.55 ± 1.08 \\
Replace With Noise & \textbf{65.07 ± 0.63} & \textbf{65.85 ± 0.64} & 64.1 ± 1.13 & \textbf{65.91 ± 1.03} & \textbf{65.7 ± 0.78} & \textbf{64.89 ± 1.28} & \textbf{66.39 ± 1.66} & \textbf{65.09 ± 0.43} \\
\bottomrule
\end{tabular}
\end{adjustbox}
\end{center}
\end{table*}

To understand the effectiveness of our method as the number of sources grows, we extended the results in Table \ref{table:synthetic_results} by significantly increasing the number of unique sources. Here, we use the low capacity CNN presented in Appendix \ref{sec:experiment_information}, with all other settings kept the same as in Table \ref{table:synthetic_results}, except that the number of epochs is reduced from $40$ to $25$ as the model architecture is smaller.

In this experiment, we see that \algsn{} performs well across the different source sizes, demonstrating its robustness as the number of sources increases. In fact, \algsn{}-\num{50000} corresponds to an experiment where the size of each source is equal to $1$ -- the standard noisy datasetting. It is interesting to see that in this case \algsn{} often performs as well or better than the baselines, suggesting its usefulness in a setting for which it was not originally designed.

These experiments take approximately $12$ hours to complete when using the compute described in Appendix \ref{sec:software}.

\subsection{Source distribution in GoEmotions dataset}
\label{sec:source_distribution_goemotions}

The GoEmotions dataset~\citep{demszky2020goemotions} was chosen for its imbalanced source distributions in the real world. Here, the training set contains source sizes in the range of $1$ to $9320$ with a mean size of $1676$ and a standard deviation of $1477$, enabling us to study the robustness of \algsn{} to imbalances in the source sizes and label distributions.

\begin{figure*}[ht]
     \centering
     \begin{subfigure}[b]{0.49\textwidth}
         \centering
         \includegraphics[width=\textwidth]{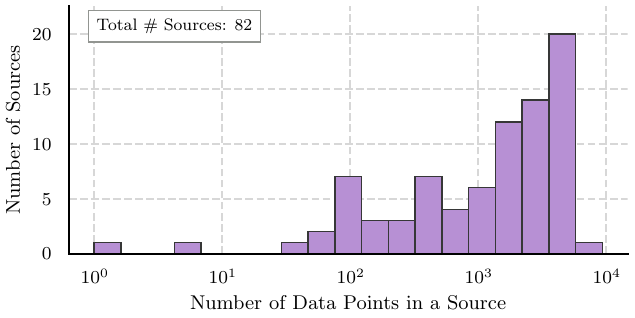}
         \caption{\textbf{The distribution of source sizes.}}
         \label{fig:goemotions_source_sizes}
     \end{subfigure}
     \begin{subfigure}[b]{0.49\textwidth}
         \centering
         \includegraphics[width=\textwidth]{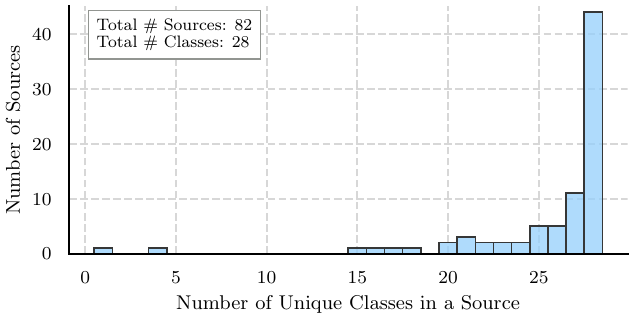}
         \caption{\textbf{Distribution of unique classes in sources.}}
         \label{fig:goemotions_source_class_unique_number}
     \end{subfigure}
    \caption{\textbf{Source and class distributions in GoEmotions.} In \subref{fig:goemotions_source_sizes} we present the number of sources with a given size, plotted on log-scale. This demonstrates the imbalance in the size of the sources in the dataset, with some containing thousands of data points, whilst others contain just a few hundred. In \subref{fig:goemotions_source_class_unique_number} we additionally explore the number of classes within each data source, showing that around half of all sources do not contain all classes, with two sources containing less than \num{5} classes.}
    \label{fig:goemotions_source_class_distribution}
\end{figure*}

In Figure \ref{fig:goemotions_source_class_distribution}, we show the distribution of source sizes and the number of classes within each source. 

Figure \ref{fig:goemotions_source_sizes} demonstrates the imbalance in the size of each source, with some sources containing thousands of data points, whilst others contain hundreds or tens of data points. A dataset with this construction allows us to test the robustness of \algsn{} to settings with uneven source sizes, a realistic situation in real-world data collection.

Furthermore, Figure \ref{fig:goemotions_source_class_unique_number} shows the distribution of the number of unique classes contained within each source. In particular, we observe that around half of all sources do not contain all the classes, with two sources containing fewer than \num{5} classes. The imbalance in the class distribution across sources is more apparent when we list the largest three classes for some of the sources.
\begin{itemize}
    \item Source 1 with size \num{37}: class 28 = $40.54\%$, class 19 = $8.11\%$, class 26 = $5.41\%$
    \item Source 2 with size \num{1137}: class 28 = $38.43\%$, class 19 = $10.82\%$, class 2 = $5.89\%$
    \item Source 4 with size \num{2702}: class 5 = $17.84\%$, class 28 = $11.84\%$, class 11 = $8.62\%$
    \item Source 8 with size \num{1253}: class 28 = $14.45\%$, class 23 = $12.05\%$, class 21 = $10.45\%$
    \item Source 16 with size \num{3224}: class 5 = $13.43\%$, class 23 = $12.38\%$, class 1 = $10.92\%$
    \item Source 32 with size \num{175}: class 28 = $50.86\%$, class 8 = $18.86\%$, class 2 = $6.86\%$
    \item Source 64 with size \num{623}: class 28 = $18.46\%$, class 4 = $11.88\%$, class 5 = $6.58\%$
\end{itemize}
These values demonstrate that sources contain large variations in the number of data points and the distributions of classes they contain. By testing with this dataset we can verify \algsn{}'s improved performance in settings in which the class distribution is uneven across sources, which may affect the source log-likelihood during training and strain the assumption we made that non-noisy sources contain similar data distributions.

\subsection{Straining the assumptions of our method}
\label{sec:straining_assumptions}

To demonstrate the robustness of \algsn{}, we now construct an experiment designed to strain the assumptions we made when introducing our method. 

To work as intended, our proposed method assumes that all non-noisy sources contain similar data distributions. This allows us to say that the weighted likelihood ratio between a source under inspection and the other sources indicates whether to increase or decrease our reliability score (Equation \ref{eq:lap_method_equation}). However, in real-world use cases, some sources might only produce a single class or a more challenging subset of classes. In such a case, we would like to understand whether our method can correctly identify the noisy sources without mistakenly labelling the more challenging data source as noisy.

To construct this setting, we combined data from both MNIST~\citep{lecun1998mnist} and CIFAR-10~\cite{krizhevsky2009learning} (Appendix \ref{sec:dataset_further_information}). First, we evenly split the MNIST data into $99$ sources and applied random label noise to $95$ of them. This created a setting with many noisy sources, which we hypothesised would be a challenging setting to learn in. We then randomly chose $2$ classes from CIFAR-10 and assigned them to a single source, giving us $99$ MNIST sources (with $95$ of them noisy) and a single CIFAR-10 source (which remained non-noisy). We then kept the same non-noisy class labelling for MNIST from the original dataset (i.e.: in the non-noisy data, a handwritten digit $0$ was assigned class $0$, etc.) and mapped the two classes from CIFAR-10 to new classes: $10$ and $11$ -- giving $12$ classes in total. Since CIFAR-10 is more challenging to classify than MNIST, this allowed us to construct a dataset in which almost all sources are easier to classify than the final source, which contains $2$ classes of more difficult data to separate. We also ran the same experiment with no random label noise applied to the MNIST sources.

To ensure that the CIFAR-10 and MNIST images are the same size and have the same number of filters, we applied a greyscale and resizing transformation to the CIFAR-10 data. We then use a simple CNN consisting of two convolutional layers with 32 and 64 filters, respectively, ReLU activation functions, max pooling layers (with kernel size $2$), and two lineaer layers mapping representations to sizes $9216$, $128$, and finally $13$. We use dropout, cross-entropy loss, and the Adam optimiser with learning rate $0.001$ and batches of size $128$, trained for $25$ epochs. The \algsn{} parameters used were $(\delta, H, \lambda) = (1.0, 50, 4.0)$ for the original data experiment and $(0.1, 50, 1.0)$ for the random label experiment. These were chosen using $5$ runs on a validation set.

\begin{table}[ht]
\begin{center}
\caption{\textbf{Difficult data results.} Mean ± standard deviation of maximum test accuracy (\%) over $5$ repeats of standard training and \algsn{} on a combination of MNIST and CIFAR-10 data with noisy labels. Here, $95$ out of $100$ sources are 100\% noisy.}
\label{table:difficult_data_results}
\scriptsize
\begin{tabular}{ccc}
\cmidrule(lr){2-3}
 & \multicolumn{2}{c}{Model Types} \\
\toprule
Noise Type & Standard & LAP (Ours) \\
\midrule
Original Data & \textbf{98.38 ± 0.4} & \textbf{98.36 ± 0.39} \\
Random Label & 67.43 ± 2.83 & \textbf{96.29 ± 0.83} \\
\bottomrule
\end{tabular}
\end{center}
\end{table}

Table \ref{table:difficult_data_results} shows the surprising results of this experiment. Since this data is made up of significant noise, it causes a large degradation in the accuracy of a standard training model when random labelling is applied to the data. However, \algsn{} allows almost all of this loss in accuracy to remain, showing significantly higher performance on the test set than the standard training method, while allowing for comparative accuracy in the absence of noise. Here, we observe that even in a contrived experiment designed as a failure case for our proposed method, \algsn{} still produces increased accuracy over this baseline. 

\begin{table}[ht]
\begin{center}
\caption{\textbf{Difficult data results on the CIFAR-10 classes.} Mean ± standard deviation of maximum test accuracy (\%) on the CIFAR-10 classes over $5$ repeats of standard training and \algsn{} trained on a combination of MNIST and CIFAR-10 data with noisy labels. This is the accuracy on just the CIFAR-10 test set, which is a subset of the test set corresponding to Table \ref{table:difficult_data_results}.}
\label{table:difficult_data_cifar_results}
\scriptsize
\begin{tabular}{ccc}
\cmidrule(lr){2-3}
 & \multicolumn{2}{c}{Model Types} \\
\toprule
Noise Type & Standard & LAP (Ours) \\
\midrule
Original Data & \textbf{0.97 ± 0.01} & 0.95 ± 0.06 \\
Random Label & \textbf{0.98 ± 0.01} & \textbf{0.97 ± 0.02} \\
\bottomrule
\end{tabular}
\end{center}
\end{table}

Furthermore, Table \ref{table:difficult_data_cifar_results} shows that we do see a small reduction in performance on the more challenging source (containing the CIFAR-10 data). On the original data, in all five runs, \algsn{} labelled the source containing the CIFAR-10 data as non-noisy as desired. However, when random label noise is applied to MNIST sources, the source containing CIFAR-10 data was incorrectly considered noisy in two of the five runs. In these two cases, the validation accuracy was lower than in the other three runs:
\begin{itemize}
    \item CIFAR-10 was incorrectly considered noisy: $23.50\%$ and $24.39\%$.
    \item CIFAR-10 was correctly considered non-noisy: $25.46\%$, $25.19\%$, $25.53\%$.
\end{itemize}
Note that the validation accuracy is low because it contains the noisy labels as well as the clean labels. Therefore, with more runs and an ensemble of the higher validation accuracy runs, higher accuracy would likely be achieved on the test set and, in particular, on the CIFAR-10 source. 

In the extreme case presented here, which violates the assumptions we made during Section \ref{sec:methods}, we find that our method is robust to challenging source class distributions and achieves markedly greater accuracy on the test set under noisy learning, whilst achieving comparative accuracy on the single challenging (but non-noisy) CIFAR-10 source.

\begin{figure*}[ht]
     \centering
     \begin{subfigure}[b]{0.32\textwidth}
         \centering
         \includegraphics[width=\textwidth]{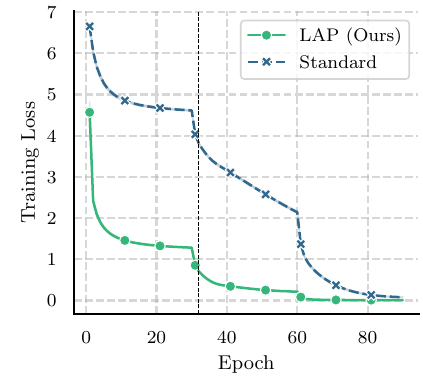}
         \caption{\textbf{Training cross entropy loss.}}\label{fig:imagenet64_random_label_and_noise_train_loss}
     \end{subfigure}
     \begin{subfigure}[b]{0.32\textwidth}
         \centering
         \includegraphics[width=\textwidth]{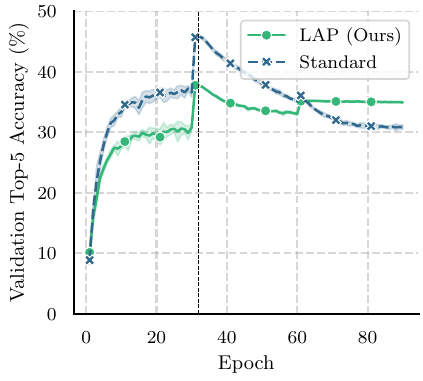}
         \caption{\textbf{Validation top-5 accuracy.}}
         \label{fig:imagenet64_random_label_and_noise_val_acc}
     \end{subfigure}
     \begin{subfigure}[b]{0.32\textwidth}
         \centering
         \includegraphics[width=\textwidth]{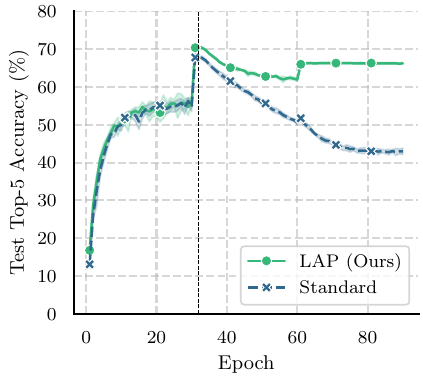}
         \caption{\textbf{Test top-5 accuracy.}}
         \label{fig:imagenet64_random_label_and_noise_top5acc}
     \end{subfigure}
    \caption{\textbf{Performance on Imagenet.} In \subref{fig:imagenet64_random_label_and_noise_train_loss} we show the cross-entropy loss on the training data at each epoch where the loss using \algsn{} represents the tempered cross-entropy loss.  
    In \subref{fig:imagenet64_random_label_and_noise_val_acc} we present the validation top-5 accuracy at each epoch.
    In \subref{fig:imagenet64_random_label_and_noise_top5acc} we present the test top-5 accuracy at each epoch to demonstrate the fitting to the noise that occurs when not using \algsn{} during late training.
    Here, $5$ out of a total of $10$ sources are $100\%$ noisy, with $2$ containing input noise, and $3$ containing label noise. The two steps in performance occur at the points at which we scale the learning rate using a scheduler (epoch $30$ and $60$). The lines and error bands represent the mean and standard deviation over the $5$ repeats. The vertical black dashed line represents the epoch at which \algsn{} achieved maximum top-5 accuracy on the test set.}
    \label{fig:imagenet_training_and_testing_curves}
\end{figure*}

\subsection{Tiny-Imagenet results}
\label{sec:tiny_imagenet_results_random_label_results}

\begin{table}[ht]
\begin{center}
\caption{\textbf{Tiny Imagenet results.} Mean ± standard deviation of maximum test top-5 accuracy (\%) over $5$ repeats of standard training and \algsn{} on Tiny Imagenet data with noisy labels. Here, $40$ out of $100$ sources are 100\% noisy.}
\label{table:tiny_imagenet_results_random_label}
\scriptsize
\begin{tabular}{ccc}
\cmidrule(lr){2-3}
 & \multicolumn{2}{c}{Model Types} \\
\toprule
Noise Type & Standard & LAP (Ours) \\
\midrule
Original Data & \textbf{61.32 ± 0.61} & 60.62 ± 0.71 \\
Random Label & 46.27 ± 1.04 & \textbf{54.48 ± 0.57} \\
\bottomrule
\end{tabular}
\end{center}
\end{table}

In an effort to demonstrate the potential of \algsn{} further, we evaluated our method on Tiny-Imagenet \citep{deng2009imagenet}, a subset of Imagenet that contains \num{100000} images from $200$ classes. In this experiment, we wanted to test the use of \algsn{} on larger images with larger numbers of sources. Firstly, we load the training and testing sets and randomly split the training set into $100$ sources uniformly. We then choose $40$ of the sources to be $100\%$ unreliable and introduce noise through random labelling. The dataset is then split into the ratio $9:1$ to produce a validation set.
For this experiment, we train a ResNet 50 architecture using the baseline training method provided for Imagenet in Pytorch~\footnote{\label{footnote:imagenet_training_script}\url{https://pytorch.org/blog/how-to-train-state-of-the-art-models-using-torchvision-latest-primitives/}}. This trains the model for $90$ epochs with a batch size of $256$ using stochastic gradient descent with an initial learning rate of $0.1$, momentum of $0.9$, and weight decay of $0.0001$, as well as a learning rate scheduler that multiplies the learning rate by $0.1$ every $30$ epochs. A version of this model (with the same parameters) is also trained using \algsn{} with $H=25$, $\delta=1.0$, and $\lambda=0.8$. As before, data is randomly assigned to mini-batches, with each batch containing multiple sources.

The results of this are available in Table \ref{table:tiny_imagenet_results_random_label} and again demonstrate the expected performance improvement when using \algsn{} for datasets with sources of unknown noise, and the maintenance of performance when data is non-noisy.

These experiments take approximately $14$ hours to complete when using the compute described in Appendix \ref{sec:software}.

\subsection{Imagenet late training test accuracy}
\label{sec:late_training}

\begin{table}[ht]
\begin{center}
\caption{\textbf{Imagenet late training results.} Mean ± standard deviation of test top-5 accuracy (\%) over the last $10$ epochs of standard training and \algsn{} on Imagenet data with noisy labels and noisy inputs, repeated $5$ times. Here, $5$ out of $10$ sources are 100\% noisy, with three of them containing label noise and 2 containing input noise.}
\label{table:imagenet_late_training_results}
\scriptsize
\begin{tabular}{ccc}
\cmidrule(lr){2-3}
 & \multicolumn{2}{c}{Model Types} \\
\toprule
Noise Type & Standard & LAP (Ours) \\
\midrule
Input and Label Noise & 42.96 ± 0.64 & \textbf{66.26 ± 0.21} \\
\bottomrule
\end{tabular}
\end{center}
\end{table}

The noisy data literature often presents model performance results on the last $x$ epochs, since in reality (as we do not have access to clean test labels) there is no way of knowing when to stop training and reduce overfitting to noisy data. In our work, in an effort to be most fair to the standard training baseline, we present the maximum performance over all of training to ensure that we are not presenting results in which the standard training baseline has significantly overfit to the noisy data. 

However, in Table \ref{table:imagenet_late_training_results} we also present the average test accuracy over the last $10$ epochs, as often reported in the literature. These results show the significant improvement that the use of \algsn{} can have on model training when it is unknown if the standard training method is overfitting to noisy data. Similarly, Figure \ref{fig:imagenet_training_and_testing_curves} presents the training, validation, and testing performance at each epoch during the training of a neural network using the standard method and \algsn{}. We can see that the training curves appear as usual (Figure \ref{fig:imagenet64_random_label_and_noise_train_loss}, with \algsn{}'s training curve representing the tempered loss) and that the validation accuracy is greater for standard training (since the validation set contains noisy labels), but that when tested on the clean labels (Figure \ref{fig:imagenet64_random_label_and_noise_top5acc}), it is clear that using \algsn{} enables considerably more robustness to noisy data. Early stopping could be used here to achieve the maximally achieving models (since the epoch of maximum accuracy on the test set is the same as the validation set), which is why we find it more informative to report the maximum test accuracy as presented in all other experiments.

\subsection{California Housing: Regression results}

\label{sec:california_housing}

\begin{table}[ht]
\begin{center}
\caption{\textbf{Standard and \algsn{} results on a regression task.} Mean ± standard deviation of minimum mean squared error of standard training and \algsn{} over $5$ repeats on the California Housing dataset with different types of noise. Here, $4$ out of $10$ sources are 100\% noisy. Note that ``Random Label" is the only noise type with a significant difference between methods. Here, smaller is better.}
\label{table:california_housing_results}
\scriptsize
\begin{tabular}{ccc}
\cmidrule(lr){2-3}
 & \multicolumn{2}{c}{Model Types} \\
\toprule
Noise Type & Standard & \algsn{} (Ours) \\
\midrule
Original Data & \textbf{0.44 ± 0.02} & \textbf{0.44 ± 0.02} \\
Random Label & 0.62 ± 0.02 & \textbf{0.45 ± 0.02} \\
\bottomrule
\end{tabular}
\end{center}
\end{table}

In Table \ref{table:california_housing_results} we present the results of using \algsn{} and standard training on a regression dataset in which we have $10$ total sources and $4$ that are $100\%$ noisy, with labels replaced with uniform noise (sampled between the minimum and maximum value of the label in the training set).

To test our method against standard training, we evaluate both on a dataset of 
\num{20640} samples of house values in California districts, with the goal to predict the median house value from the U.S census data for that region -- a regression task. We randomly divide the data into training and testing with a $8:2$ ratio for each of the $5$ runs we performed over each noise type. We train a multilayer perceptron as described in Section \ref{sec:experiment_information} for $200$ epochs with a batch size of $256$, leaning rate of $0.001$ using stochastic gradient descent with momentum of $0.9$ and weight decay of $0.0001$. A version of this model (with the same parameters) is also trained using \algsn{} with $H=25$, $\delta=1.0$, and $\lambda=0.8$. As before, data is randomly assigned to mini-batches, with each batch containing multiple sources. This was repeated $5$ times for each setting.

Table \ref{table:california_housing_results} shows that \algsn{} improves the performance on the test set significantly over the baseline for random labelling noise and maintains the performance when no noise is present. This is as expected and again illustrates our method's effectiveness, but this time on a regression task.

These experiments take approximately $2$ hours to complete when using the compute described in Appendix \ref{sec:software}.

\subsection{Additional Compute Example}

\label{sec:compute_example}

In this section, we briefly describe the additional cost of applying \algsn{} with a specific example.

In practise using our implementation, with an MLP with hidden sizes $20-2000-2000-5$, and batches of $512$ samples (with 20 features, 5 classes, and data generated from 10 sources) the forward-backward pass through the model (without source reweighting) takes 15300 $\mu$ s ± 164 $\mu$ s and the reweighting of losses takes 469 $\mu$ s ± 10.8 $\mu$ s, increasing the time by ~3\%. With 50 and 100 sources, the reweighting of losses takes 1000 $\mu$ s ± 11.3 $\mu$ s and 1990 $\mu$ s ± 49.2 $\mu$ s respectively. 

Using the same set-up, with batches of 2048 in size and 10 sources, the forward-backward pass (without reweighting) takes 41000 $\mu$ s ± 689 $\mu$ s whilst the loss reweighting takes 467 $\mu$ s ± 8.51 $\mu$ s (no significant change as the batch size increases), increasing the time to compute the loss and gradients on a batch by ~1\%. 

We believe that this is a fair trade-off for the potential improvement in performance.

\end{document}